\documentclass[11pt]{article} 
\usepackage{url}
\usepackage{smile}
\usepackage{graphicx} 
\usepackage{algorithm}
\usepackage{algorithmic}
\usepackage{todonotes}
\usepackage{epstopdf}
\usepackage{wrapfig}
\usepackage[colorlinks, linkcolor=blue, anchorcolor=blue, citecolor=blue]{hyperref}
\usepackage[margin=1in]{geometry}
\usepackage[normalem]{ulem}
\usepackage[export]{adjustbox}
\usepackage{mathtools, cuted}
\usepackage{dsfont}
\usepackage{natbib}
\usepackage{enumerate}
\usepackage{enumitem}
\newcommand{\sGamma}{\Gamma_{\mathrm{sym}}}

\newcommand{\id}{\mathrm{Id}}
\newcommand{\idu}{\mathrm{Id}_{x^*}}

\newcommand{\lbr}[1]{\left\langle#1\right\rangle}

\newcommand{\fnorm}[1]{\|#1\|_{\rm{F}}}
\newcommand{\tnorm}[1]{\|#1\|_{2}}
\begin{document}
\title{\huge \bf Noise Regularizes Over-parameterized Rank One Matrix Recovery, Provably}

\author
{Tianyi Liu, Yan Li, Enlu Zhou, and Tuo Zhao\thanks{Tianyi Liu is now affiliated with ByteDance. The work is done when Tianyi Liu is a Ph.D. student at Georgia Tech; Yan Li, Enlu Zhou, and Tuo Zhao are affiliated with Georgia Tech;  Email: tianyi.liu@bytedance.com, tourzhao@gatech.edu.}}

\date{}

\maketitle
\begin{abstract}

We investigate the role of noise in optimization algorithms for learning over-parameterized models. Specifically, we consider the recovery of a rank one matrix $Y^*\in\RR^{d\times d}$ from a noisy observation $Y$ using an over-parameterization model.  We parameterize the rank one matrix $Y^*$ by $XX^\top$, where $X\in\RR^{d\times d}$. We then show that under mild conditions, the estimator, obtained by the randomly perturbed gradient descent algorithm using the square loss function, attains a mean square error of $\cO(\sigma^2/d)$, where $\sigma^2$ is the variance of the observational noise. In contrast, the estimator obtained by gradient descent without random perturbation only attains a mean square error of $\cO(\sigma^2)$. Our result partially justifies the implicit regularization effect of noise when learning over-parameterized models, and provides new understanding of training over-parameterized neural networks.
\end{abstract}

\section{Introduction}

Deep neural networks have revolutionized many research areas, and achieved the state-of-the-art performance in many computer vision \citep{krizhevsky2012imagenet, goodfellow2014generative, Long_2015_CVPR}, natural language processing \citep{graves2013speech, bahdanau2014neural, young2018recent} and signal processing tasks \citep{yu2010deep}. Such huge successes cannot be well explained by conventional wisdom. These deep neural networks are significantly over-parameterized -- using more parameters than statistically necessary. However, training these neural networks does not require explicit regularization or constraints to control the model complexity.

There have been two major lines of theoretical research on demystifying the over-parameterization phenomenon. One line of research attempts to investigate the training of deep neural networks from a pure optimization perspective. \citet{liang2018adding,Sharifnassab2020Bounds,liang2019revisiting} show that under properly simplified settings, the over-parameterization can eliminate spurious local optima of the training objective, and all obtained local optima become global. Therefore, over-parameterization makes the optimization landscape benign, which eases the training of neural network. However, these results are not relevant to the generalization performance of neural nets. 

Another line of research attempts to connects the deep neural networks to reproducing kernel functions. \citet{du2018gradient, jacot2018neural,allen2018convergence,arora2019fine} show that under certain conditions, training the over-parameterized neural networks by gradient descent is equivalent to training a kernel machine, which is often referred to as Neural Tangent Kernel (NTK) in existing literature. Therefore, adding more neurons only makes the behavior of deep neural networks behave more close to that of their corresponding reproducing kernel functions. By further exploiting such a connection, they show that the global optima of the training objective can be obtained by the gradient descent (GD) algorithm, However, as shown in \citet{weinan2020comparative}, these results cannot explain the generalization performance well, as the equivalent reproducing kernel functions still suffer from the curse of dimensionality.

Complementary to the aforementioned two lines of research, there have been some empirical investigations on the role of noise in optimization algorithms for training over-parameterized neural networks. For example, \citet{keskar2016large} show that the stochastic gradient descent (SGD) algorithms with small batch sizes yield significantly better generalization performance than those with large batch sizes. This clearly indicates that the noise plays a very important role on implicitly controlling the model complexity of over-parameterized neural networks. Unfortunately, due to the complex structures of deep neural networks and current technical limit, establishing theory for understanding the noise in SGD is very challenging. Though some of the aforementioned work consider SGD, they only consider small learning rates and large batch sizes to make the noise negligible such that its behavior is close to GD. Hence, their results cannot justify the advantage of SGD for training over-parameterized models.

To flesh out our understanding the role of noise for training over-parameterized models, we propose to analyze a simpler but nontrivial alternative problem -- over-parameterized matrix factorization using perturbed gradient descent (P-GD). Specifically, we consider the recovery of a symmetric rank one matrix $Y^*\in\RR^{d\times d}$ from its noisy observation $Y$ under over-parameterization model. Different from existing work, which usually parameterizes $Y^*$ as the outer product of two vectors, we factorize $Y^*$ as the product of two matrices $XX^\top$, where $X\in\RR^{d\times d}$. Therefore, we are essentially using $d^2$ parameters rather than statistically necessary $d$ parameters. To recover $Y^*$, we solve the following  optimization problem,
\begin{align}\label{overall-objective}
\min_{X\in\RR^{d\times d}}\frac{1}{4}\norm{Y-XX^\top}_{\rm F}^2.
\end{align}
We then solve \eqref{overall-objective} using a perturbed form of gradient descent P-GD, which injects independent noise to iterates, and then evaluates gradient at the perturbed iterates. Note that our algorithm is different from SGD in terms of the noise. For our algorithm, we inject independent noise to the iterate $X_t$ and use the gradient evaluated at the perturbed iterates. The noise of SGD, in contrast, usually comes from the training sample. As a consequence, the noise of SGD has very complex dependence on the iterate, which is difficult to analyze. 

We further analyze the computational and statistical properties of the P-GD algorithm. Specifically, at the early stage, noise helps the algorithm to avoid regions with undesired landscape, including saddle points. After entering the region with benign landscape, the noise induces an implicit regularization effect, and P-GD eventually converges to an estimator $\hat{X}$, which attains a mean square error of  $\cO(\sigma^2/d)$ with overwhelming probability, i.e.,
\begin{align*}
\frac{1}{d^2}\norm{\hat{X}\hat{X}^\top-Y^*}_{\rm F}^2 = \cO_P\Big(\frac{\sigma^2}{d}\Big),
\end{align*}
where $\sigma^2$ is the variance of the observational noise. For comparison, if we solve \eqref{overall-objective} by GD without random perturbation, and the obtained estimator only attains a mean square error of $\cO(\sigma^2)$. To the best of our knowledge, this is the first theoretical result towards understanding the role of noise in training over-parameterized models.

Our work is closely related to \citet{li2017algorithmic}, which analyze GD for solving over-parameterized matrix sensing problem. Specifically, they show that when initialized at a sufficiently small magnitude, GD also has an implicit regularization effect and can approximately recover low rank matrix under the RIP condition. Their theory, however, only works for noiseless cases. Our theory complements their results under the noisy setting, and demonstrates that the noise of algorithms can also contribute to the implicit regularization effects for training over-parameterized models. 

The rest of the paper is organized as follows: Section \ref{sec_model} introduces the rank-1 matrix factorization problem and the perturbed gradient descent algorithm to solve it. Section \ref{sec_convergence} presents the main theorem showing that P-GD converges to solutions with smaller mean square error than GD. Section \ref{sec_numerical} verifies our theoretical result numerically on rank-1 matrix recovery, rank-$r$ matrix recovery and also rectangular matrix recovery. The discussion on the extension of our theoretical results and also related literature is presented in Section \ref{sec_discussion}.

{\bf Notations:} Let $\cS$ be a subspace of $\RR^d,$ we use $\mathrm{Proj}_{\cS}(\cdot)$ to denote the projection of a vector or matrix to $\cS.$ For a vector  $v\in \RR^d$ and matrix $A\in \RR^{d\times d},$ we use $\id_{v}A$ to denote the projection of each column of $A$ onto the subspace  $\mathrm{span}(v)=\{x\in \RR^d| x=\alpha v, \alpha\in \RR\}.$ $\id$ is the identity matrix. The  ball with radius $r$ in $\RR^d$ and its sphere are denoted as $\mathds{B}(1)$ and $\SSS(1),$ respectively. For matrices $A,B\in \RR^{n\times m},$  we use $\inner{A}{B}$ to denote the Frobenius inner product, i.e., $\inner{A}{B}=\tr(A^\top B).$ $\norm{A}_{\rm{F}}$ and $\tnorm{A}$ denotes the Frobenius norm and spectral norm of $A,$ respectively. 

\section {Model and Algorithm}\label{sec_model}
We first describe the over-parameterized rank one matrix factorization problem. Specifically, we observe a matrix $Y\in\RR^{d\times d}$, where
\begin{align*}
Y = Y^*+ \Gamma,
\end{align*}
where $Y\in\RR^{d\times d}$ is an unknown rank one matrix, and $\Gamma\in\RR^{d\times d}$ is a random noise matrix with each entry i.i.d. sampled from some sub-Gaussian distribution with $\EE \Gamma_{ij} = 0$ and $\EE\Gamma_{ij}^2=\sigma^2$. We recover $Y^*$ by solving the following problem:
\begin{align}\label{rankr_mf}
\hat{X} = \argmin_{X \in \RR^{d\times d}} \cF(X),~\textrm{where}~\cF(X) =\frac{1}{4}  \norm{XX^\top - Y}_{\rm{F}}^2.
\end{align}  
The estimator of $Y^*$ can be obtained by $\hat{Y}=\hat{X}\hat{X}^\top$. Here $\hat{Y}$ is over-parameterized with $d^2$ parameters in $\hat{X}$, while the intrinsic dimension of the rank one matrix $Y^*$ is only $d$. We do not use any explicit regularizer to control the search space of $X$.

We then describe the perturbed gradient descent (P-GD) algorithm for solving \eqref{rankr_mf}. Specifically, at the $(t+1)$-th iteration, we first inject a random noise matrix $W_t\in\RR^{d\times d}$ to $X_t$, 
\begin{align*}
\tilde{X}_{t}&=X_t+W_t,
\end{align*} 
where each column of $W_t$ is independently sampled from $\mathrm{UNIF}(\SSS(\nu))$, and $\SSS(\nu)$ denotes the hypersphere with radius $\nu$ centered at 0. Note that we have $\fnorm{W_t}^2=d\nu^2$. We then update $X_t$ using the gradient of $\cF(X)$ at $\tilde{X}_t$,
\begin{align}\label{pgd}
X_{t+1} = X_t -\eta\nabla\cF(\tilde{X}_{t})=X_t -\eta\left(\tilde{X}_t\tilde{X}_{t}^\top-Y_{\rm sym}\right)\tilde{X}_t,
\end{align}
where $Y_{\rm sym} = (Y+Y^\top)/2$. The P-GD algorithm is essentially solving the following stochastic optimization problem,
\begin{align}\label{mat_fact_eq}
\min_{X\in\RR^{d\times d}}\tilde\cF(X)=\EE_{W}\cF(X+W),
\end{align}
where each column of $W$ is independently sampled from $\mathrm{UNIF}(\SSS(\nu))$. Note that \eqref{mat_fact_eq} can be viewed as a smooth approximation of \eqref{rankr_mf} by convolution using a uniform kernel. The smoothing effect further induces implicit regularization effect to the estimator. 

The P-GD algorithm is also related to the randomized smoothing in existing literature. It was first proposed by \citet{duchi2012randomized} to handle convex non-smooth optimization. \citet{zhou2019towards,jin2017escape,lu2019pa} further show that the random perturbation can also help escape from saddle points and spurious optima.



\section{Convergence Analysis}\label{sec_convergence}
We study the convergence properties of our proposed perturbed gradient descent (P-GD) algorithm. 
Before presenting our main results, we first introduce the subspace dissipative condition, which is frequently used in our proof and is defined as follows. 
\begin{definition}[Subspace Dissipativity]\label{SC}
	Let $\cS$ be a subspace of $\RR^d$ and $x_\cS=\mathrm{Proj}_\cS(x)$ be the projection of $\forall x\in\RR^d$ into $\cS.$ 
	For any operator  $\cH:~ \RR^d\rightarrow \RR^d,$ we say that $\cH$ is $(c_{\cS},\gamma_{\cS},\cS)$-subspace dissipative with respect to (w.r.t.) the subset $\cX^*\subseteq \RR^d$ over the set $\cX\supseteq\cX^*$, if for every $x\in\cX$, there exist an $x^*\in\cX^*$ and two positive universal constants $c_\cS$ and $\gamma_\cS$ such that
	\begin{align}\label{con_sc}
	\left\langle\mathrm{Proj}_\cS(\cH(x)),x_\cS-x^*_\cS\right\rangle\geq c_\cS \norm{x_\cS-x^*_\cS}_2^2-\gamma_{\cS}. 
		\end{align}
	Here, $\cX$ is called the subspace dissipative region of the operator $\cH$.	
\end{definition}
The intuition behind the subspace dissipative condition is that $\mathrm{Proj}_\cS(\cH(x))$ has a  positive fraction pointing towards $x_S^*$ up to certain perturbation. When the algorithm iterates along  $-\cH(x)$, its projection in $\cS$  can gradually evolve towards $x_S^*$ and finally converge to a neighborhood of  $x_S^*.$ 


We then introduce two assumptions on the signal noise ratio and the initialization, respectively. Specifically, the first assumption requires noise $\Gamma$ not to overwhelm the ground truth low rank matrix $Y^*$. For notational simplicity, we denote $Y^*=x^*x^{*\top}$.
\begin{assumption}[Signal-Noise-Ratio]\label{ass_noise} 
There exist some universal constants $C_0, C_1$ such that
\begin{align}
&\norm{x^*}_2 \geq C_0,\quad\sigma\leq \frac{C_1}{d},\quad\fnorm{\sGamma}\leq 2d\sigma,\quad\max\{\tnorm{\sGamma x^*},\tnorm{\sGamma}\}\leq C_1\sqrt{d}\sigma,
\end{align}
where $$\sGamma = Y_{\rm sym}-Y^* = (Y+Y^\top)/2-Y^* = (\Gamma+\Gamma^{\top})/2.$$
\end{assumption}
The spectral and Frobenius norms of the noise $\sGamma$ are of order $\cO(1/\sqrt{d})$ and $\cO(1)$, respectively, while $x^*$ is non-degenerate and yields a sufficiently large signal noise ratio. 

Note that \citet{li2019symmetry} show that 0 is a strict saddle point to \eqref{rankr_mf}. The second assumption requires the initialization of the P-GD algorithm to be sufficiently distant from $0$. 
\begin{assumption}[Proper Initialization]\label{ass_initial} 	
	$X_0$ is bounded and sufficiently away from $0,$ i.e.,
	\begin{align}\label{eq_initial}
\fnorm{X_0}^2\leq 1- C_1\sqrt{d\sigma^2}, \tnorm{X_0^\top x^*}^2\geq C_1^2d\sigma^2.
	\end{align}	
\end{assumption}
Assumption \ref{ass_initial} can be further relaxed to an arbitrary initialization within a hyperball centered at $0$. We do not consider such a relaxation, since it is not directly related to the regularization effect of the noise, but makes the convergence analysis much more involved. 

\begin{remark}
Note that both Assumptions \ref{ass_noise} and \ref{ass_initial} are deterministic. Later in Lemmas \ref{lem_noise} and \ref{lem_initial}, we will show that both assumptions hold with high probability, given that $\Gamma$ is sub-Gaussian and our initialization is random within a ball.
\end{remark}

We then present our main results in the following theorem.
\begin{theorem}[Convergence Rate of P-GD]\label{thm_main}
Suppose that Assumptions \ref{ass_noise} and \ref{ass_initial} hold for $\Gamma$ and $X_0$, respectively.  For any $\delta\in(0,1),$ we choose 
\begin{align*}
\nu^2=C_1 \sqrt{d\sigma^2}\quad\textrm{and}\quad\eta\leq \eta_0=\cO\Big(\frac{\sigma^2}{d^2}\Big(\log \frac{1}{\delta}\Big)^{-1}\Big).
\end{align*}
Then there exists some generic constant $c_0$ such that with probability at least $1-\delta$, we have
\begin{align*}
\frac{1}{d^2}	\fnorm{X_tX_t^\top-Y^*}^2\leq c_0 \frac{\sigma^2}{d},
\end{align*}
for all t's such that $\tau_{0}\leq t\leq T={\cO}(\eta^{-2})$, where $$\tau_{0}={\cO}\Big(\frac{1}{\eta }\log\frac{1}{d\sigma^2}\log
\frac{1}{\delta}\Big).$$
\end{theorem}
Theorem \ref{thm_main} implies that the noise plays an important role on regularizing the over-parameterized model during training, and induces a bias towards low complexity estimators. The estimation error is optimal for noisy rank one matrix factorization. For comparison, we can invoke the theoretical analyses in \citet{jain2015global} and show that GD does not have such a regularization effect and converges to a solution denoted by $X_{\rm GD}$, where $X_{\rm GD}X_{\rm GD}^\top$ is essentially the positive semidefinite approximation of $Y_{\rm sym}$. Therefore, $X_{\rm GD}$ only attains a suboptimal estimation error,
\begin{align*}
\frac{1}{d^2}\fnorm{X_{\rm GD}X_{\rm GD}^\top-Y^*}^2 = \cO(\sigma^2).
\end{align*}
As can be seen, P-GD outperforms GD in terms of mean square error for recovering the underlying low rank matrix $Y^*$ by a factor of $d$.


The proof of Theorem \ref{thm_main}  is very involved. Due to the space limit, we only present a proof sketch here. Please see more details in Appendix \ref{proof}.
\begin{proof}[{\bf \emph{Proof Sketch}}]
Without loss of generality, we assume $\norm{x^*}_2=1.$ We start with a meta proof plan.
Specifically, we decompose $X_t$ into its projections in the subspace spanned by $x^*$ and its orthogonal complement  as follows: 
\begin{align}\label{decomp}
X_t={ \idu X_t}+{(\id-\idu)X_t}={ x^*r_t^{\top}}+E_t,
\end{align}
where $r_t=X_t^\top x^*.$   Note that the signal term $R_t=x^*r_t^{\top}$ always satisfies $R_tR_t^\top=\tnorm{r_t}^2 Y^*,$ which is a multiple of the ground truth matrix. Therefore, any solution satisfying $\tnorm{r_t}^2=1$ and $E_t=0$ gives the exact recovery.  
In light of this fact, we show that P-GD can find a solution such that $\tnorm{r_t}$ is approximately $1$ and $E_t$ stays small. To facilitate our analysis, we write down the update of $r_t$ and $E_t$ as follows.
\begin{align*}
r_{t+1}&=X_{t+1}^\top x^*=r_t-\eta   \nabla_X\cF(X_t+W_t)^\top x^*,\\
E_{t+1}&=(\id-\idu) X_{t+1}=E_t-\eta (\id-\idu) \nabla_X\cF(X_t+W_t).
\end{align*}
We further denote  the gradient of $\cF$ with respect to $r$ and $E$ as $\nabla_{r}\cF(X)=\nabla_X\cF(X)^\top x^*$ and $\nabla_{E}\cF(X)= (\id-\idu)\nabla_X\cF(X),$ where  $r=X^\top x^*, $ and $E= (\id-\idu) X.$ The next lemma shows that $\nabla_{r}\cF$ and $\nabla_{E}\cF$ satisfy the subspace dissipative condition.
\begin{lemma}[Subspace Dissipativity]\label{lem_pd}
	For any $X\in\RR^{d\times d},$ $\nabla_E\cF$ satisfies
	\begin{align}\label{pd_E}
	\lbr{\EE_W[\nabla_E \cF(X+W)],E}\geq  \left((2d+1)\frac{\nu^2}{d}-\tnorm{\sGamma}\right)\fnorm{E}^2-\frac{1}{4}\tnorm{\sGamma x^{*}}^2.
	\end{align}
Let $a=1-(2d+1)\frac{\nu^2}{d}+x^{*\top}\sGamma x^*,$ then  $\nabla_{r}\cF$ satisfies  the  inequality below  if $\tnorm{r}^2\geq a.$
	\begin{align}\label{pd_r}
	\lbr{\EE_W[  \nabla_{r}\cF(X+W)],r}\geq \tnorm{r}^2(\tnorm{r}^2-a)-\frac{1}{4}\fnorm{\sGamma u_*}^2.
	\end{align}
	Moreover, when $\fnorm{E}^2\leq c^2\tnorm{\sGamma x^*}, \tnorm{\sGamma x^*}^2\leq \tnorm{r}^2\leq a,$ for some constant $c>0,$   then  $-\nabla_{r}\cF$ satisfies the following inequality. 
		\begin{align}\label{pd_r2}
	\lbr{\EE_W[-  \nabla_{r} \cF(X+W)],r}\geq\tnorm{r}^2(a-\tnorm{r}^2) -(c^2+c)\fnorm{\sGamma u_*}^2.
	\end{align}
	\end{lemma}

Lemma \ref{lem_pd} helps describe the converge pattern of $E_t$ and $r_t.$ Specifically, the subspace dissipativity holds for $\nabla_{E}\cF(X+W)$ globally, which implies that the orthogonal part $E_t$ vanishes independent of $r_t$. The convergence of $\tnorm{r_t}^2,$ however, is more complicated. On the one hand, \eqref{pd_r} suggests that when $\tnorm{r_t}^2$  exceeds $a,$ P-GD tends to decrease  the norm $\tnorm{r_t}^2.$ On the other, when $\tnorm{r_t}^2$ is small, \eqref{pd_r2} suggests $\tnorm{r_t}^2$ will increase to $a$ only after $\fnorm{E_t}^2$ is sufficiently small. Combining these two aspects, $\tnorm{r_t}^2$ will move towards and stay close to $a\approx 1.$  We remark that \eqref{pd_r2} requires $\fnorm{E_t}^2$ to be small. Therefore, the convergence of $\tnorm{r_t}^2$ happens after that of $\fnorm{E_t}^2.$

Before showing the convergence, we provide a  lemma showing that the trajectory of P-GD is bounded with high probability. This lemma helps us bound high order terms in the proof.
 \begin{lemma}[Boundedness of Trajectory]\label{lem_bound}
 	Suppose $\Gamma$ and $X_0$ satisfy  Assumptions \ref{ass_noise} and \ref{ass_initial}, respectively. For any $\delta\in(0,1),$ we choose $\nu^2=C_1 \sqrt{d\sigma^2}$ and $$\eta\leq \eta_1=\min\left\{\cO\left(\frac{1}{d}\left(\log \frac{1}{\delta}\right)^{-1}\right),\cO\left(\frac{1}{d^2}\right)\right\}.$$ Then with probability at least $1-\delta,$  for  $t\leq T=\cO(\frac{1}{\eta^2}),$  $$\fnorm{X_t}^2\leq  4d.$$
 \end{lemma}
  Following our discussions, we first show the convergence of  $\fnorm{E_t}^2$ in the next lemma.
\begin{lemma}[Convergence of $E_t$]\label{lem_E_converge}
Suppose $\Gamma$ and $X_0$ satisfy  Assumptions \ref{ass_noise} and \ref{ass_initial}, respectively.   For any $\delta\in(0,1),$ we choose $\nu^2=C_1 \sqrt{d\sigma^2}$ and $$\eta\leq \eta_2=\min\left\{\cO\left(\frac{\sigma}{d^3}\left(\log \frac{1}{\delta}\right)^{-1}\right),\cO\left(\frac{\sigma^2}{d^2}\right)\right\},$$ then  with probability at least $1-\delta,$ 
\begin{align}\label{eq_e1}
\fnorm{E_t}^2\leq \fnorm{E_0}^2+ c_1\sqrt{d{\sigma}^2}
\end{align}
 holds for all t's such that $t\leq T={\cO}(\eta^{-2})$, and 
\begin{align}\label{eq_e2}
 \fnorm{E_t}^2\leq  c_1\sqrt{d{\sigma}^2} 
\end{align}

holds for all t's such that $\tau_{1}\leq t\leq T={O}(\eta^{-2})$
 where
	$c_{1}$ is an absolute constant and $$\tau_{1}={\cO}\Big(\frac{1}{\eta \sqrt{d\sigma^2}}\log\frac{1}{{d\sigma^2}}\log
	\frac{1}{\delta}\Big).$$
\end{lemma}

In addition to the convergence result \eqref{eq_e2}, the boundedness of $\fnorm{E_t}^2$ in \eqref{eq_e1} will help us show that $\tnorm{r_t}^2$ always stays away from the strict saddle point $0$  as shown in the following lemma.	\begin{lemma}[Avoid Strict Saddle]\label{lem_away}
Suppose $\Gamma$ and $X_0$ satisfy  Assumptions \ref{ass_noise} and \ref{ass_initial}, respectively.  Assume \eqref{eq_e1} holds for all $t>0.$ For any $\delta\in(0,1),$ we choose $\nu^2=C_1 \sqrt{d\sigma^2}$ and $$\eta\leq \eta_3=\min\left\{\cO\left(\sigma^4\left(\log \frac{1}{\delta}\right)^{-1}\right),\cO\left(\frac{\sigma^2}{d^2}\right)\right\},$$ we then have with probability at least $1-\delta,$  for all $t\leq \cO(1/\eta^2),$
\begin{align}\label{eq_r1}
\tnorm{r_{t}}^2\geq \tnorm{\sGamma x^*}^2.
\end{align}
\end{lemma}
Given that $\fnorm{E_t}^2$ becomes sufficiently small in Lemma \ref{lem_E_converge}, and $r_t$ stays distant from zero, we can then invoke subspace dissipative condition \eqref{pd_r} and \eqref{pd_r2} and show that $\tnorm{r_t}^2$ will converge to $1$ in the following lemma.

\begin{lemma}[Convergence of $r_t$]\label{lem_r_converge}
Suppose $\Gamma$ satisfies  Assumption \ref{ass_noise}. Assume \eqref{eq_e2} and \eqref{eq_r1} hold for all $t>0.$ For any $\delta\in(0,1),$ we choose $\nu^2=C_1 \sqrt{d\sigma^2}$ and $$\eta\leq \eta_4=\min\left\{\cO\left(\sigma^4\left(\log \frac{1}{\delta}\right)^{-1}\right),\cO\left(\frac{\sigma^2}{d^2}\right)\right\}.$$ Then  with probability at least $1-{\delta},$  
\begin{align}\label{eq_r2}
|\tnorm{r_t}^2-1|\leq c_2\sqrt{d\sigma^2}
\end{align}
 for all t's such that $\tau_{2}\leq t\leq T={O}(\eta^{-2})$, where
	$c_{2}$ is an absolute constant and $$\tau_{2}={\cO}\Big(\frac{1}{\eta}\log\frac{1}{d\sigma^2}\log
	\frac{1}{\delta}\Big).$$
\end{lemma}

Note that the recovering error can be rewritten as follows.
\begin{align}
\fnorm{X_tX_t^\top-Y^*}^2&=(1-\tnorm{r_t}^2)^2+2\tnorm{E_tr_t}^2+\fnorm{E_tE_t^\top}^2\nonumber\\
&\leq (1-\tnorm{r_t}^2)^2+2\fnorm{E_t}^2\tnorm{r_t}^2+\fnorm{E_t}^4.\label {err1}
\end{align}
Combining \eqref{eq_e2} and \eqref{eq_r2}, we know that P-GD has already entered and stays in the region with small recovery error. We remark that a naive treatment of the cross term $\tnorm{E_t r_t}^2$ as in \eqref{err1}  will result in a recovery error $\cO(\sqrt{d\sigma^2})$ that dominates \eqref{err1}, with a  worse dependency on $d$. Instead, we take a  more refined approach to bound the cross term by  analyzing its optimization trajectory.
\begin{lemma}[Convergence of $E_tr_t$]\label{lem_er_converge}
Suppose $\Gamma$ satisfies  Assumption \ref{ass_noise}. Assume  \eqref{eq_e2} and \eqref{eq_r1} hold for all $t>0.$ For any $\delta\in(0,1),$ we choose $\nu^2=C_1 \sqrt{d\sigma^2}$ and $$\eta\leq \eta_5=\cO\left(d\sigma^2\left(\log \frac{1}{\delta}\right)^{-1}\right).$$ Then  with probability at least $1-{\delta}{},$ 
 $$\tnorm{E_tr_t}^2\leq c_3{d\sigma^2}$$  holds  for all t's such that $\tau_{3}\leq t\leq T={\cO}(\eta^{-2})$, where
	$c_{3}>0$ is an absolute constant and $$\tau_{3}={\cO}\Big(\frac{1}{\eta}\log\frac{1}{d\sigma^2}\log
	\frac{1}{\delta}\Big).$$
\end{lemma}
The proof of Lemmas \ref{lem_pd}–\ref{lem_er_converge} requires supermartingale-based analysis, which is very involved and technical.  See more details in Section \ref{sec_super1} and Appendix \ref{proof}.

Finally, using the conclusions  of Lemmas \ref{lem_E_converge}, \ref{lem_r_converge} and \ref{lem_er_converge}, we have with high probability that
\begin{align*}
\fnorm{X_tX_t^\top-Y^*}^2&=(1-\tnorm{r_t}^2)^2+2\tnorm{E_tr_t}^2+\fnorm{E_tE_t^\top}^2\\&\leq ( c_2^2+2c_2+c_1^2) d \sigma^2
\end{align*}
holds when $ \tau_1+\tau_2+\tau_3\leq t\leq T.$
Take $c_0=c_1^2+c_2^2+2c_3, \tau_0=\tau_1+\tau_2+\tau_3$ and
\begin{align*}
\eta \leq \eta_0 &=\cO\left(\frac{\sigma^2}{d^2}\left(\log \frac{1}{\delta}\right)^{-1}\right)\leq \min\{\eta_1, \eta_2,\eta_3,
\eta_4, \eta_5\},
\end{align*}
where the last inequality holds since $\sigma = \cO(1/d),$ and we prove that
\begin{align*}
\frac{1}{d^2}\fnorm{X_tX_t^\top-Y^*}^2\leq c_0 \frac{\sigma^2}{d},
\end{align*}
holds with high probability for all $ \tau_0\leq t\leq T.$
\end{proof}

Next, we verify that the noise matrix and the initialization of P-GD satisfy Assumptions \ref{ass_noise} and \ref{ass_initial}, respectively. For noise matrix $\Gamma$ with i.i.d sub-Gaussian entries, Assumption \eqref{ass_noise} holds with high probability by applying the standard concentration result as  in the next lemma.
\begin{lemma}[Signal-Noise-Ratio]\label{lem_noise}
	For any $\delta\in (0,1),$ with high probability at least $1-\delta$, we have 
	\begin{align}	
	&\max\{	\tnorm{\sGamma x^*}, \tnorm{\sGamma}\}\leq C\sqrt{d}\sigma+C\sigma\sqrt{\log\frac{8}{\delta}},\nonumber\\&~~\fnorm{\sGamma}\leq d\sigma+\sigma\sqrt{2C\log\frac{8}{\delta}},
	\end{align} 
	where $C$ is some absolute constant. Moreover, take $\delta=\cO(\exp(-d))$ and we have 
	 	\begin{align*}	
		\max\{	\tnorm{\sGamma x^*}, \tnorm{\sGamma}\}\leq C_1\sqrt{d}\sigma,~~\fnorm{\sGamma}\leq 2d\sigma.
	 \end{align*} 
\end{lemma}

Furthermore, P-GD with random initialization in a unit ball satisfies Assumption  \ref{ass_initial} with high probability as shown in the next lemma.
\begin{lemma}[Proper Initialization]\label{lem_initial}
 Given a random initialization $X_0=x_0x_0^\top$ where $x_0\sim \mathrm{UNIF}(\mathds{B}(1)),$ then with  probability at least $1-\cO\left(d^{-\frac{1}{4}}\right),$
 \begin{align*}
 \fnorm{X_0}^2\leq 1- C_1\sqrt{d\sigma^2}, \tnorm{X_0^\top x^*}^2\geq C^2d\sigma^2.
 \end{align*}	
\end{lemma}

\subsection{Super-martingale Theorem}\label{sec_super1}
In this section, we briefly introduce the key technique behind our analysis.
We first provide a super-martingale based theorem frequently used in our ensuing analysis of perturbed gradient descent algorithm. 
Such a theorem can also be adapted to analyzing other stochastic recursive algorithm satisfying certain conditions, and hence can be of independent interest.
\begin{theorem}\label{thm_general}
Given a random sequence $\{x_t\}\in \RR^d$ 
 satisfying $x_{t+1}=x_t-\eta f(x_t, \xi_t), \quad\forall t\geq0, $
 where $x_0\in \RR^d$ is known,  $f$ is some bounded real valued function and $\xi_t\in \RR^d$ represents the randomness in the update. Let $g$ be some real valued function on $\RR^d.$ If there exist constants $\alpha_0\in \RR,\beta>0, \lambda>0, \phi>0$ such that  $(1-\eta\beta)^{-t}(g(x_t)-\alpha_0-\eta\lambda)$ is a super-martingale  for any $\eta>0$ satisfying  $\eta\beta<1, $  i.e.,
 \begin{align}\label{eq_dyn}
\EE[(1-\eta\beta)^{-t-1}(g(x_{t+1})-\alpha_0-\eta\lambda)|\cF_t]
\leq (1-\eta\beta)^{-t}(g(x_t)-\alpha_0-\eta\lambda),
\end{align}
where $\cF_t=\sigma\{x_\tau,\tau\leq t\}.$Then for any $\delta\in[0,1],\alpha\geq \alpha_0,$ we have the following conclusions.

\noindent {\bf Part I.} If we take $\displaystyle\eta\leq \min\left\{\frac{\alpha_0}{2(\max f)\cI_{g(x_0)>2\alpha}},\frac{\alpha_0}{4\lambda}\right\}.$ With probability at least $1-\delta,$ there exists $t\leq \tau',$ such that
	$$g(x_t)\leq 2\alpha,$$
	where \begin{align*}
	\tau'=\left\{
	\begin{array}{c l}	
	\cO\left( \frac{1}{\eta \beta}\log\frac{4(g(x_0)-\frac{5}{4}\alpha_0)}{\alpha}\log\frac{1}{\delta}\right), & g(x_0)>2\alpha;\\
	0, & \mathrm{o.w.}
	\end{array}\right.
	\end{align*}
	
\noindent {\bf Part II.}	
Moreover, if we further have 
\begin{align}\label{eq_diff}
|g(x_{t+1})-\EE[g(x_{t+1})|\cF_t]|\cI_{\{g(x_t)\leq 4\alpha\}}\leq \phi\eta,
\end{align}
and  take 
\begin{align*}
\eta\leq \min\Bigg\{&\cO\left(\frac{\alpha^2\beta}{\phi^2}\left(\log \frac{1}{\delta}\right)^{-1}\right),~\frac{\alpha_0}{2(\max f)\cI_{g(x_0)>2\alpha}},~\frac{\alpha_0}{4\lambda}\Bigg\},
\end{align*}
 then with probability at least $1-\delta,$
$$g(x_t)\leq 4\alpha,$$
for any $\tau'\leq t\leq T=\cO(1/\eta^2).$
\end{theorem}
Theorem \ref{thm_general} has two parts of results. Part I states that when the update satisfies \eqref{eq_dyn}, with properly chosen step size, the sequence $\{x_t\}$ can enter the region where $g$ is bounded by a pre-specified constant $\alpha$ in polynomial time. Part II ensures that  the sequence will stay in this region for long enough time. 
Please refer to Section \ref{super_martingale} for the detailed proof.

To utilize Theorem  \ref{thm_general} in analyzing our P-GD algorithm, we only need to check whether  $g(x) = \norm{x-x^*}_2^2$ meets the condition stated in \eqref{eq_dyn}. In fact, when the subspace dissipativity condition \eqref{con_sc} is satisfied, we have
\begin{align*}
\EE\left[\norm{x_{t+1}-x^*}_2^2|\cF_t\right] =& \norm{x_{t}-x^*}_2^2 - 2 \eta \EE\left[ \left\langle f(x_t, \xi_t),x_t-x^*_\cS\right\rangle|\cF_t\right]+ \eta^2\EE\left[ \norm{ f(x_t, \xi_t)}_2^2|\cF_t\right]\\
\leq& (1-2\eta c_S)\norm{x_{t}-x^*}_2^2 + 2\eta(\gamma_S + \cO(\eta))
\end{align*} 
where $c_S, \gamma_S$ are the constants of subspace dissipativity conditions. By simple manipulation, the above inequality can be shown to be equivalent to the following inequality
\begin{align*}
&\EE\left[(1-2\eta c_S)^{-t-1}\left(\norm{x_{t+1}-x^*}_2^2 -\left(\frac{\gamma_S}{c_S} + \cO(\eta)\right) \right)|\cF_t\right]\\
&~~~~~~~~~~~~~~~~~~~~~~~~~~~\leq (1-2\eta c_S)^{-t}\left(\norm{x_{t}-x^*}_2^2 - \left(\frac{\gamma_S}{c_S} + \cO(\eta)\right)\right),
\end{align*} 
which is in the same form as in \eqref{eq_dyn}. 
Therefore, by exploiting subspace dissipativity in conjunction with our developed  super-martingale theorem, we are poised to 
prove the key elements Lemma  \ref{lem_pd}–\ref{lem_er_converge} in the analysis of P-GD.
\section{Numerical Experiments}\label{sec_numerical}

In this section, we demonstrate the regularization effect of noise using numerical experiments. Specifically, we compare  P-GD algorithm with gradient descent (GD) with small and large initialization to show that noise induces  a bias towards low complexity estimators.

\subsection{Noisy Positive Semidefinite Matrix Recovery}

We consider recovering a positive semidefinite (PSD)  matrix $Y^* = X^* X^{*\top}$, with $X^* \in \RR^{d \times r}$. We first present experiments for $r=1$ to support our theory, then we conduct experiments on $r = 3$ and show that the general rank-r PSD matrix recovery exhibits similar behavior.
\noindent {\bf Rank-1.} We set the ground truth matrix  $Y^*=x^*x^{*\top}$, where $x^*=(1,1,\cdots, 1)\in \RR^{d},$ and $d=30.$  
The noise matrix $\Gamma$ has i.i.d. Gaussian entries with mean $0$ and variance $\sigma^2=0.1.$ 
  We run P-GD, GD with small and large initializations   to solve \eqref{rankr_mf}. Specifically, GD-Small is initialized  at $\frac{1}{d}A,$  where $A$ is a random orthogonal matrix as suggested in \cite{li2017algorithmic}, while GD-Large is initialized  at $\frac{1}{\sqrt{d}}B,$ where $B$ is a random  matrix with i.i.d. standard normal entries. P-GD takes the same initialization as GD-Large. In every iteration of P-GD , we perturb the iterate  with $\Gamma$ having i.i.d. $\mathrm{Unif(\SSS(\nu))}$ columns, where  $\nu^2={0.4\sqrt{d\sigma^2}}.$ All three algorithms are run with  $\eta=0.25\sigma^2/d^2=2.7\times 10^{-6}$ for $T=1\times 10^8$ iterations. 
  \begin{figure*}[t]
\centering
\includegraphics[width=0.95\linewidth]{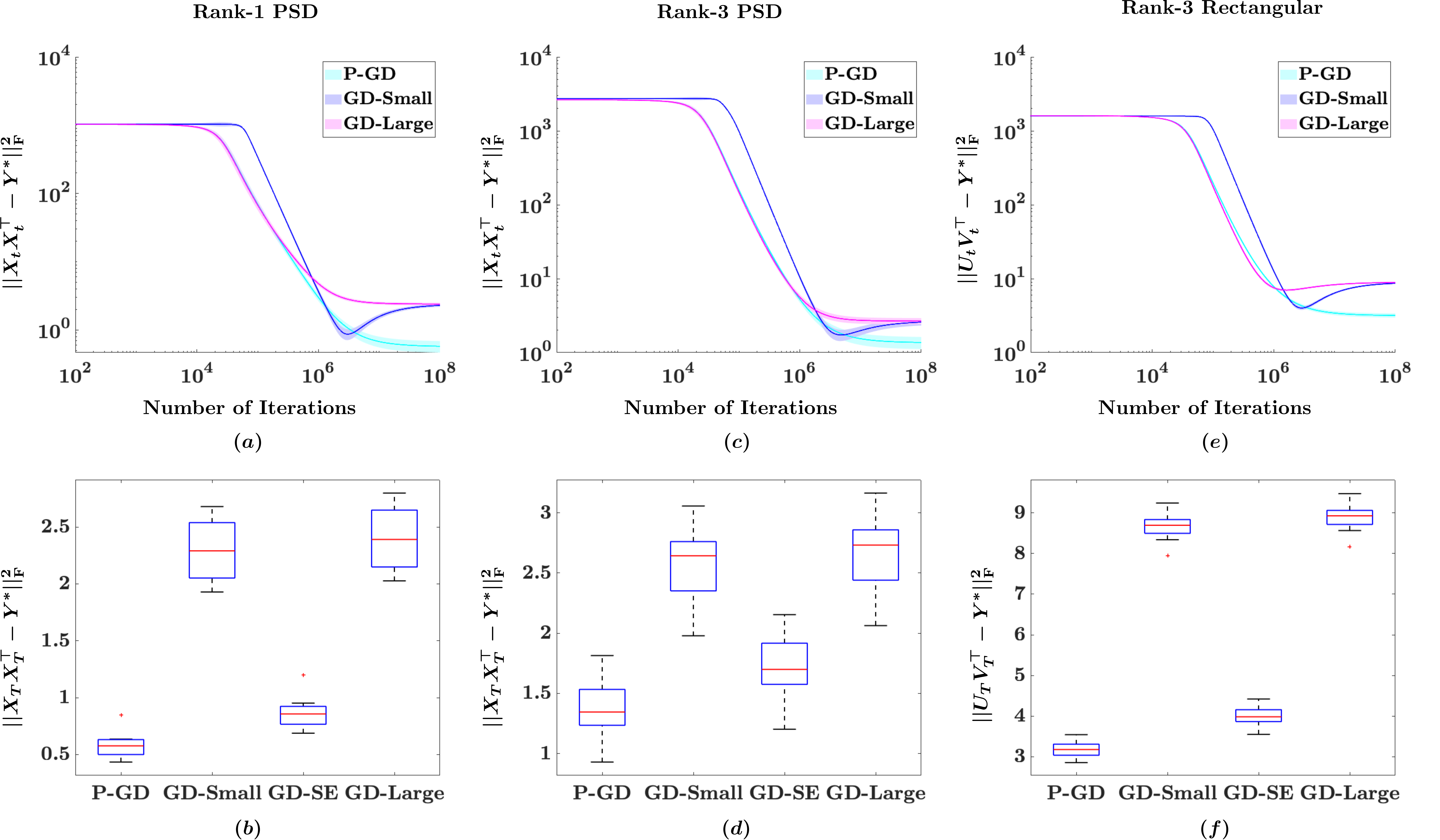}
\caption{ Average learning curves and final recovery error box plots of P-GD,  GD with small initialization (GD-Small) and with large initialization (GD-Large). X,Y-axes are in log scale. The band in (a), (c), (e) represents standard deviation. (a)-(b): Rank-1 positive semidefinite matrix recovery.  (c)-(d): Rank-3 positive semidefinite matrix recovery. (e)-(f): Rank-3 rectangular matrix recovery. GD-Small shows a regularization effect in the early stage but overfits later;  P-GD performs the best. GD-Small with Early Stopping (GD-SE) achieves significantly better recovery error than other GD's, but still worse than P-GD.}
\label{fig:rank1}	
\end{figure*}

  The results of $20$ repeated runs are summarized in Figure \ref{fig:rank1}.(a) and (b).  The average learning curve in Figure \ref{fig:rank1}.(a) shows that the convergence of GD with small initialization  has two phases. Specifically, it first iterates towards the low complexity solutions and achieve a recovery  error $0.9$ in around $3\times10^6$ iterations,  which is consistent with the algorithmic regularization effect of GD  shown in \cite{li2017algorithmic}. In the second stage, however, GD-small overfits the observational noise and finally attains a larger recovery error about $2$, which is similar to GD-Large.   In Figure \ref{fig:rank1}.(b), we plot the final recovery error of the three algorithms. We also plot the minimal recovery error obtained by GD-Small  and name it GD-Small with  early stopping (GD-SE).  It can be seen GD-SE avoids overfitting and can obtain a significantly lower recovery error than GD-Small. This observation justifies the regularization effect of early stopping in gradient descent learning. However, GD-SE still performs worse than P-GD. Different from GD, P-GD always converges to the estimators with lower recovery error around $0.5$, even with large initialization.  This suggests that noise induces implicit bias towards the low complexity solutions in training over-paramterized models. 
  
\noindent {\bf Rank-3.} We then consider rank-3 PSD matrix recovery. We choose  set $Y^*=X^*X^{*\top}$, and $X^*\in\RR^{d\times 3}$ with i.i.d. standard Gaussian entries. One can verify that $Y^*$ is a rank-3 PSD matrix with probability 1. We choose other experiment settings the same as those of the rank-1 case  except $\nu^2={0.25\sqrt{d\sigma^2}}.$ 
The results of $20$ repeated runs are summarized in Figure \ref{fig:rank1}.(c) and (d).  
We have similar observations like in the rank-1 case
from the learning curve and boxplot  of GD and P-GD. 
\subsection{Noisy Rectangular Matrix Recovery}
 We perform experiments on rectangular matrix factorization,  to show that the regularization effect of noise is not limited to symmetric matrix factorization problems.
We set ground truth matrix $Y^*=U^* V^{*\top}$, where $U^*\in \RR^{d\times 3}$ and $V^*\in \RR^{d\times 3}.$  We recover $Y^*$ by solving the following over-parameterized nonconvex optimization problem.
\begin{align}\label{rec_mf}
\left(\hat{U}, \hat{V}\right)= \argmin_{U \in \RR^{d\times d}, V \in \RR^{d\times d}} \frac{1}{2}  \norm{UV^\top - Y}_{\rm{F}}^2,
\end{align} 
where $Y=Y^*+\Gamma$ is a noisy observation of $Y^*.$ P-GD  solving \eqref{rec_mf} takes the following update:
\begin{align}\label{rec_pgd}
U_{t+1}&=U_t-\eta \nabla_U \cF(U_t+W_t, V_t+Z_t),\nonumber
\\V_{t+1}&=V_t-\eta \nabla_V \cF(U_t+W_t, V_t+Z_t),
\end{align}
where $\nabla_U \cF(U, V)=(UV^\top-Y)V,~~~\nabla_V \cF(U, V)=(UV^\top-Y)^\top U.$ Note that without perturbation, GD will converge exactly to an optimal solution such that $U_tV_t=Y.$
In our experiment, we choose $d=30$ and $U^*, V^*$  to be two random rectangular matrices with i.i.d. standard Gaussian entries. The noise matrix $\Gamma$ has i.i.d. Gaussian entries with mean $0$ and variance $\sigma^2=0.1.$ We run P-GD, GD-Small and GD-large to solve \eqref{rec_mf}.  GD-Small is initialized  at $\frac{1}{d}(A_1, A_2),$  where $A_1, A_2$ are a random orthogonal matrix, while GD-Large is initialized  at $\frac{1}{\sqrt{d}}(B_1,B_2),$ where $B_1, B_2$ is a random  matrix with i.i.d. standard normal entries. P-GD takes the same initialization as GD-Large. We run P-GD with perturbation noise $W_t,Z_t$ taking i.i.d. $\mathrm{Unif(\SSS(\nu))}$ columns, where  $\nu^2={0.6\sqrt{d\sigma^2}}.$ All three algorithms are run with  $\eta=0.25\sigma^2/d^2=2.7\times 10^{-6}$ for $T=1\times 10^8$ iterations. 

The results of $20$ repeated experiments are summarized in Figure \ref{fig:rank1}.(e) and (f).  
We observe similar phenomenon of GD and P-GD as that in the rank-1 PSD matrix recovery, which advocates that the regularization effect of noise appears in general rectangular matrix recovery.


\section{Discussions}\label{sec_discussion}
{\bf Extension to Rank-r Matrix Recovery: } We can  extend our theoretical analysis to rank-r PSD matrix recovery. 
Similar to the projection in \eqref{decomp}, we project each iterate into 
 the subspaces spanned by each eigenvectors of $Y^*$, and the orthogonal complement. The subspace dissipative conditions of each subspace can be obtained following similar lines to the proof of Lemma \ref{lem_pd}.  We can then apply our super-martingale type analysis and show that P-GD can achieve the optimal convergence rate $\cO(\frac{r\sigma^2}{d})$ for the rank-r case under some conditions on the eigen-value. The analysis, however, will be much more involved. We believe our results on  the rank-1 case has already unveiled the regularization effect of noise and left technical extensions as our future work. 
  
\noindent{\bf Extension to Rectangular Matrix Recovery: }  Our theoretical analysis can potentially extend to rectangular matrix recovery \eqref{rec_mf} by reducing the problem to symmetric PSD matrices as in \citet{ge2017no}. Denote $W_t^\top=(U_t^\top, V_t^\top)$ and 
$W^{*\top}=(U^{*\top}, V^{*\top}),$ where $Y^*=U^* V^{*\top}.$ One can verify $N^*=W^*W^{*\top}$ is a PSD matrix. Recovering $Y^*$ by P-GD \eqref{rec_pgd} can be viewed as recovering $N^*$  by applying P-GD on $W_t.$ The problem is then reduced to  rank-r PSD matrix recovery. To complete the analysis, we need theoretical guarantees on the equivalence of this reduction in our noisy observation case, which is left for future research.

\noindent{\bf Biased Stochastic Gradient Approximation: } In our P-GD algorithm,  the random perturbation to the iterates makes the gradient approximation biased. We remark that the biased stochastic gradient approximation also appears in training neural networks. Specifically, neural nets are often trained by SGD combining with many  regularization techniques such as batch normalization (BN), weight decay, dropout and etc. These tricks help overcome overfitting. Meanwhile, since they essentially change the network structure or the loss function, the stochastic gradient  in SGD becomes biased with respect to the original objective \citep{helmbold2015inductive,helmbold2017surprising,mianjy2018implicit,luo2018towards}. Such a biased approximation is worth further investigation to unveil their importance  in learning over-parameterized models.

\noindent{\bf Regularization Effect: } Our theoretical results provide new insights towards understanding the regularization effect of SGD in training deep neural networks. Specifically, besides the algorithmic regularization induced by deterministic first order algorithms (such as GD as shown in \citet{li2017algorithmic}), our theory suggests that noise also plays an important role in regularizing over-parameterized models.

\noindent{\bf Related Literature:}  \citet{blanc2020implicit} also study the implicit regularization for SGD type of algorithms based on a different problem setup, i.e., a 2-layer neural network without over-parameterization. They consider noise perturbation on labels instead of on parameters as in our paper and do not provide any explicit recovery error bound. Moreover,  \citet{blanc2020implicit}  consider regularizing the $l_2$ norm of gradient, which is equivalent to adding a regularizer defined by $||(XX^\top-Y)X||_F^2$ in our setting.   To our best knowledge, there is no existing literature that shows this regularizer help find solutions with low complexity. 

Some other papers study related problems but have fundamental differences with our work.  \citet{haochen2020shape} consider perturbing labels while our work consider perturbing parameters. \citet{du2018power} consider a problem without the underlying low complexity generating models, while we take advantage of an underlining low rank generating model and provide an estimation error bound analysis. Most importantly, we study the implicit regularization effect of noise without any explicit regularizers used in their work.



\bibliography{ref}
\bibliographystyle{apalike}

\appendix
\section{Proof of Theorem \ref{thm_general}}\label{super_martingale}
\begin{proof}

\noindent{\bf Part I.}	We first show that with probability at least $1-\delta,$ there exists $t\leq \tau_0$ such that $g(x_t)\leq2 \alpha.$ We only need to consider the case where $g(x_0)>2\alpha_0.$ Let $\cE_t=\{g(x_\tau)\geq2\alpha,\forall \tau\leq t\}, G_{t}=(1-\eta\beta)^{-t}(g(x_t)-\alpha_0-\eta\lambda).$ Then by \eqref{eq_dyn}, 
	we have 
	$$\EE[G_{t+1} \cI_{\cE_t}\cF_t]\leq \EE[G_{t} \cI_{\cE_t}]\leq  \EE[G_{t} \cI_{\cE_{t-1}}].$$
	The last inequality holds since when $\cI_{\cE_{t-1}}=1$ while $\cI_{\cE_{t}}=0,$ $G_t\geq 0$ when $\eta\leq \min\{\frac{\alpha_0}{2\max f},\frac{\alpha_0}{4\lambda}\}$. Then $\{G_t\cI_{\cE_{t-1}}\}$ are a supermartingale sequence. Then 
		$$\PP(\cE_t)\leq \PP(g(x_t)\geq 2\alpha)\leq \frac{\EE[g(x_t)]}{2\alpha}\leq \frac{(1-\eta\beta)^t(g(x_0)-\alpha_0-\eta\lambda)+\alpha_0+\eta\lambda }{2\alpha}\leq \frac{3}{4},$$
		when $t\geq \frac{1}{\eta \beta}\log\frac{4(g(x_0)-\frac{5}{4}\alpha_0)}{\alpha}.$ Recursively applying the above lines for $\cO(\log\frac{1}{\delta})$ times, we know there exists $t\leq \tau$ such that $g(x_t)\leq 2\alpha$ with probability at least $1-\delta,$ where 
	$$\tau_0=\cO\left( \frac{1}{\eta \beta}\log\frac{4(g(x_0)-\frac{5}{4}\alpha_0)}{\alpha}\log\frac{1}{\delta}\right).$$
	
\noindent{\bf Part II.} Then we show, with high probability, for any $\alpha\geq \alpha_0,$ $g(x_t)\leq 4\alpha$ for long enough time. Let $\cH_t=\{g(x_\tau)\leq4\alpha,\forall \tau\leq t\}.$ By \eqref{eq_dyn}, we have 
\begin{align}\label{ineq_sup}
\EE[G_{t+1} \cI_{\cH_t}\cF_t]\leq \EE[G_{t} \cI_{\cH_t}]\leq  \EE[G_{t} \cI_{\cH_{t-1}}].
\end{align} 
	Then $\{G_t\cI_{\cE_{t-1}}\}$ are a super-martingale sequence.  We then bound the difference between $G_t\cI_{\cE_{t-1}}$ and $\EE[G_t\cI_{\cE_{t-1}}|\cF_t].$	
	\begin{align}\label{ineq_diff}
d_t=|G_t\cI_{\cE_{t-1}}-\EE[G_t\cI_{\cE_{t-1}}|\cF_t]|=(1-\eta\beta)^{-t}|g(x_{t+1})-\EE[g(x_{t+1})|\cF_t])|\leq (1-\eta\beta)^{-t}\phi \eta.
	\end{align}
Denote $D_t=\sqrt{\sum_{i=0}^{t}d_i^2}$. By Azuma's Inequality, we get
$$\mathbb{P}\left(G_t\cI_{\cH_{t-1}}-G_0 \geq{\cO}\left(1\right)D_t\log^{\frac{1}{2}}\left(\frac{1}{\eta^2\delta}\right)\right)
\leq \exp\left(-\frac{{\cO}\left(1\right)D_t^2\log\left(\frac{1}{\eta^2\delta}\right)}{2\sum_{i=0}^{t}d_i^2}\right)
={\cO}\left(\eta^2\delta\right).$$

Therefore, with at least probability $1-{\cO}\left(\eta^2\delta\right)$, we have
\begin{align*}
g(x_t)
&\leq \left(1-\eta\beta\right)^t\left(g(x_0)-\alpha_0-\eta\lambda\right) 
+ {\cO}\left(1\right)\left(1-\eta\beta\right)^t D_t\log^{\frac{1}{2}}\left(\frac{1}{\eta^2\delta}\right)
+ \alpha_0+\eta\lambda\\
&\leq g(x_0)
+ {\cO}\left(1\right)\frac{\phi\eta}{\sqrt{\eta\beta}}\log^{\frac{1}{2}}\left(\frac{1}{\eta^2\delta}\right)
+\frac{5}{4}\alpha\leq 4\alpha,
\end{align*}
where the last line holds, since we can always find $\eta=\min\left\{\cO\left(\frac{\alpha^2\beta}{\phi^2}\left(\log \frac{1}{\delta}\right)^{-1}\right),\frac{\alpha_0}{4\lambda}\right\}$ to satisfy the condition.

The above inequality shows that if $\cH_t$ holds, then $\cH_{t+1}$ holds with at least probability $1-{\cO}\left(\eta^2\delta\right)$. Hence, with at least probability $1-\delta$, we have $g(x_t)\leq 4\alpha$ for all $t\leq T={\cO}\left(\frac{1}{\eta^2}\right)$.

Combine Part I and  Part 2, properly rescale $\delta,$ and we prove the theorem.
\end{proof}
\section{Proof of Technical Lemmas}\label{proof}

\subsection{Proof of Lemma \ref{lem_pd}}
\begin{proof}
For notational simplicity, we denote $\gamma= \frac{\nu}{\sqrt{d}}.$ We start from the subspace dissipative condition for $E_t.$  We will use the fact $Y^*E=0.$
\begin{align*}
&~~~~~~\lbr{E,\EE_W (\id-\idu) \nabla_X\cF(X+W)}\\&=	\lbr{E, (\id-\id_{S})\left( (XX^\top-Y)X+(2d+1)\gamma^2 X\right) }\\
&=\lbr{E,(2d+1)\gamma^2 E+ (\id-\id_{S})\left( (XX^\top-Y)X\right) }\\
&=(2d+1)\gamma^2\fnorm{E}^2+\lbr{XX^\top-Y,(\id-\id_{S})EX^\top }\\
&=(2d+1)\gamma^2\fnorm{E}^2+\lbr{XX^\top-Y,EX^\top }\\
&=(2d+1)\gamma^2\fnorm{E}^2+\lbr{XX^\top-Y^*,EX^\top }-\lbr{\sGamma,EX^\top }\\
&=(2d+1)\gamma^2\fnorm{E}^2+\lbr{XX^\top,EX^\top }-\lbr{\sGamma,EX^\top }\\
&=(2d+1)\gamma^2\fnorm{E}^2+\fnorm{EX^\top}^2-\lbr{\sGamma,EX^\top }\\
&=(2d+1)\gamma^2\fnorm{E}^2+\fnorm{EE^\top}^2+\fnorm{E Z^\top}^2-\lbr{\sGamma,EE^\top }-\lbr{\sGamma,EZ^\top }\\
&\geq ((2d+1)\gamma^2-\tnorm{\sGamma})\fnorm{E}^2+\fnorm{EZ^\top}^2-\tnorm{\sGamma }\tnorm{EZ^\top}\\
&\geq ((2d+1)\gamma^2-\tnorm{\sGamma})\fnorm{E}^2-\frac{1}{4}\tnorm{\sGamma}^2.
\end{align*}
The last inequality holds since given $b>0,$ $x^2-bx\geq -\frac{b^2}{4}$ for $\forall x>0.$ Then we obtain the inequality \eqref{pd_E}. 

 We next prove the subspace dissipative condition for $r.$ 
 \begin{align}\label{eq_inner_r}
&~~~~~\lbr{r_t, \EE\left[\nabla_X\cF(X_t+W_t)^\top x^*\right]}\nonumber\\
&=\lbr{r_t, X_t^\top (X_tX_t^\top-Y_{\mathrm{sym}})x^*+(2d+1)\gamma^2 X_t^\top x^*}\nonumber\\
 &=(2d+1)\gamma^2\norm{r_t}_2^2+\lbr{r_t, X_t^\top(X_tX_t^\top-Y^*)x^*}-\lbr{r_t, X_t^\top\sGamma x^*}.
 \end{align}
 We calculate the last two terms separately. Note that $X_t^\top(X_tX_t^\top-Y^*)x^*=X_t^\top X_t r_t-r_t=(\norm{r_t}_2^2-1)r_t+E_t^\top E_t r_t.$ Insert this equation in the second term in \eqref{eq_inner_r} and we have $$\lbr{r_t, X_t^\top(X_tX_t^\top-Y^*)x^*}=(\norm{r_t}_2^2-1)\norm{r_t}_2^2+\norm{E_tr_t}_2^2.$$ Moreover, the last term in\eqref{eq_inner_r} can be calculated as follows.
 \begin{align*}
 \lbr{r_t, X_t^\top\sGamma x^*}&=r_t^{\top} X_t^\top\sGamma x^*\\
 &=r_t^{\top} (r_t x^{*\top}+E_t^\top)\sGamma x^*\\
 &=\norm{r_t}_2^2x^{*\top}\sGamma x^*+r_t^{\top} E_t^\top\sGamma x^*.
 \end{align*}
	Let $a=1-(2d+1)\gamma^2+x^{*\top}\sGamma x^*.$ Then we have 
  \begin{align*}
 \lbr{r_t, \EE\left[\nabla_X\cF(X_t+W_t)^\top x^*\right]} &=(2d+1)\gamma^2\norm{r_t}_2^2+(\norm{r_t}_2^2-1)\norm{r_t}_2^2\\&~~~~~~~~~~~~~~~+\norm{E_tr_t}_2^2-\norm{r_t}_2^2x^{*\top}\sGamma x^*-r_t^{\top} E_t^\top\sGamma x^*\\
 &=\left(\norm{r_t}_2^2-1+(2d+1)\gamma^2-x^{*\top}\sGamma x^*\right)\norm{r_t}_2^2+\norm{E_tr_t}_2^2-r_t^{\top} E_t^\top\sGamma x^*\\
 &\geq (\norm{r_t}_2^2-a)\norm{r_t}_2^2+\norm{E_tr_t}_2^2-\tnorm{ E_tr_t}\tnorm{\sGamma x^*}\\
 &\geq  (\norm{r_t}_2^2-a)\norm{r_t}_2^2-\frac{1}{4}\tnorm{\sGamma x^*}^2.
 \end{align*}
This proves the inequality \eqref{pd_r}.
On the other hand, 
  \begin{align*}
 \lbr{r_t, -\EE\left[\nabla_X\cF(X_t+W_t)^\top x^*\right]} &=-(2d+1)\gamma^2\norm{r_t}_2^2-(\norm{r_t}_2^2-1)\norm{r_t}_2^2\\&~~~~~~~~~~~~~~~-\norm{E_tr_t}_2^2+\norm{r_t}_2^2x^{*\top}\sGamma x^*+r_t^{\top} E_t^\top\sGamma x^*\\
 &=\left(a-\norm{r_t}_2^2\right)\norm{r_t}_2^2-\norm{E_tr_t}_2^2+r_t^{\top} E_t^\top\sGamma x^*\\
 &\geq (a-\norm{r_t}_2^2)\norm{r_t}_2^2-\norm{E_tr_t}_2^2-\tnorm{ E_tr_t}\tnorm{\sGamma x^*}\\
 &\geq  (a-\norm{r_t}_2^2)\norm{r_t}_2^2-(c^2+c)\tnorm{\sGamma x^*}^2.
 \end{align*}
  We prove the inequality \eqref{pd_r2}.
\end{proof}
 \subsection{Proof of Lemma \ref{lem_bound}}
 \begin{proof}
 	Given our choice of $\nu,$ we have with probability at least $1-\delta,$ $d\gamma^2\geq \tnorm{\sGamma}.$ Given our initialization, we have $\fnorm{X_0}^2=1\leq 4d.$
 	\begin{align*}
 	\EE[\fnorm{X_{t+1}}^2|\cF_t]&=\EE\left[\fnorm{X_t-\eta \nabla_X\cF(X_t+W_t)}^2|\cF_t\right]\\
 	&=\fnorm{X_t}^2-2\eta \EE[ \lbr{X_t,\nabla_X\cF(X_t+W_t)}|\cF_t]+\eta^2\EE\left[\fnorm{\nabla_X\cF(X_t+W_t)}^2|\cF_t\right]\\
 	&=\fnorm{X_t}^2-2\eta  \lbr{X_t,\EE[\nabla_X\cF(X_t+W_t)|\cF_t] }+\eta^2\EE\left[\fnorm{\nabla_X\cF(X_t+W_t)}^2|\cF_t\right]\\
 	&=\fnorm{X_t}^2-2\eta  \lbr{X_t,\nabla_X\cF(X_t)+(2d+1)\gamma^2 X_t}+\eta^2\EE\left[\fnorm{\nabla_X\cF(X_t+W_t)}^2|\cF_t\right]\\
 	&=\fnorm{X_t}^2-2\eta  \lbr{X_t,  (X_tX_t^\top-Y^*)X_t-\sGamma X_t+(2d+1)\gamma^2 X_t}\\&\hspace{+2.5in}+\eta^2\EE\left[\fnorm{\nabla_X\cF(X_t+W_t)}^2|\cF_t\right]\\
 	&=(1-2\eta (2d+1)\gamma^2)\fnorm{X_t}^2-2\eta\fnorm{X_t^\top X_t}^2+2\eta\lbr{X_t, Y^*X_t}\\&\hspace{+1.5in}+2\eta\lbr{X_t, \sGamma X_t}+\eta^2\EE\left[\fnorm{\nabla_X\cF(X_t+W_t)}^2|\cF_t\right].
 	\end{align*}
 	Note that $$\lbr{X_t, Y^*X_t}=\tr(X_t^\top Y^*X_t)\leq \fnorm{ X_t}^2,$$ 
 	$$\fnorm{X_t^\top X_t}^2=\sum_{i,j}(X_t^\top X_t)^2_{i,j}\geq \sum_{i}(X_t^\top X_t)^2_{i,i}\geq\frac{1}{d}\left(\sum_{i}(X_t^\top X_t)_{i,i}\right)^2=\frac{1}{d}\fnorm{X_t}^4, $$
 	and  $$\lbr{X_t, \sGamma X_t}\leq \tnorm{\sGamma}\fnorm{ X_t}^2.$$ 
 	
 	Then we have 
 	\begin{align*}
 	\EE[\fnorm{X_{t+1}}^2|\cF_t]&\leq \left(1-2\eta (2d+1)\gamma^2+2\eta(1+\tnorm{\sGamma})\right)\tnorm{X_t}^2-2\eta\frac{1}{d}\fnorm{X_t}^4+\eta^2\EE[\fnorm{\nabla_X\cF(X_t+W_t)}^2].
 	\end{align*}
 	or equivalently,
 	\begin{align*}
 	&~~~~~~\EE[\fnorm{X_{t+1}}^2-d|\cF_t]\\
 	&\leq \left(1-2\eta (2d+1)\gamma^2+2\eta\tnorm{\sGamma}\right)(\fnorm{X_t}^2-d)-2\eta\frac{1}{d}\fnorm{X_t}^2(\fnorm{X_t}^2-d)\\&~~~~~~~~-2\eta d((2d+1)\gamma^2-\tnorm{\sGamma})+\eta^2\EE[\fnorm{\nabla_X\cF(X_t+W_t)}^2]\\
 	&\leq  \left(1-2\eta (2d+1)\gamma^2+2\eta\tnorm{\sGamma}-2\eta\frac{1}{d}\fnorm{X_t}^2\right)(\fnorm{X_t}^2-d)\\&~~~~~~~~-2\eta d\left((2d+1)\gamma^2-\tnorm{\sGamma}\right)+\eta^2\EE[\fnorm{\nabla_X\cF(X_t+W_t)}^2]\\
 	&\leq  \left(1-2\eta (2d+1)\gamma^2+2\eta\tnorm{\sGamma}-2\eta\frac{1}{d}\fnorm{X_t}^2\right)(\fnorm{X_t}^2-d)+\eta^2\EE[\fnorm{\nabla_X\cF(X_t+W_t)}^2]\\
 	&\leq   \left(1-2\eta (2d+1)\gamma^2+2\eta\tnorm{\sGamma}-2\eta\right)(\fnorm{X_t}^2-d)+\eta^2\EE[\fnorm{\nabla_X\cF(X_t+W_t)}^2].
 	\end{align*}
 	
 	Let $\cE_t$ be the event $\left\{\fnorm{X_\tau}^2\leq 4d,\forall \tau\leq t\right\}.$ Then 
 	\begin{align*}
 	\EE\left[(\fnorm{X_{t+1}}^2-d)\cI_{\cE_t}|\cF_t\right]&\leq   \left(1-2\eta (2d+1)\gamma^2+2\eta\tnorm{\sGamma}-2\eta\right)(\fnorm{X_t}^2-d)\cI_{\cE_t}+\\&\hspace{+2.5in}\eta^2\EE[\fnorm{\nabla_X\cF(X_t+W_t)}^2]\cI_{\cE_t}\\
 	&\leq  \left(1-2\eta (2d+1)\gamma^2+2\eta\tnorm{\sGamma}-2\eta\right)(\fnorm{X_t}^2-d)\cI_{\cE_t}+\eta^2C_1\cI_{\cE_t}.
 	\end{align*}
 	where $C_1=\cO(d^3).$ Let $\lambda_1=\frac{C_1}{2 (2d+1)\gamma^2-2\tnorm{\sGamma}+2}=\cO(d^3),  \beta_1=2 (2d+1)\gamma^2-2\tnorm{\sGamma}+2.$ 
 	Equivalently, we have the following inequality.
 	\begin{align*}
 	\EE\left[(\fnorm{X_{t+1}}^2-d-\eta \lambda_1)\cI_{\cE_t}|\cF_t\right]&\leq   \left(1-\beta_1\right)(\fnorm{X_t}^2-d-\eta \lambda_1)\cI_{\cE_t}.
 	\end{align*}
 	
 	We further denote $G_t=(1-\beta_1)^{-t}(\fnorm{X_t}^2-d-\eta\lambda_1).$ Then we have 
 	$$\EE[G_{t+1}\cI_{\cE_t}]\leq \EE[G_{t}\cI_{\cE_{t}}]\leq \EE[G_{t}\cI_{\cE_{t-1}}]. $$  
 	Then \eqref{ineq_sup} is satisfied.
 	We then bound the difference between  $G_{t}\cI_{\cE_{t-1}} $ and the conditional expectation $\EE[G_{t}\cI_{\cE_{t-1}}|\cF_{t-1}]. $
 	\begin{align*}
 	d_t&=\big| G_{t}\cI_{\cE_{t-1}}-\EE[G_{t}\cI_{\cE_{t-1}}|\cF_{t-1}] \big|\\
 	&=(1-\beta_1)^{-t}\big|2\eta\left( \EE[ \lbr{X_{t-1},\nabla_X\cF(X_{t-1}+W_{t-1})}|\cF_{t-1}]- \lbr{X_{t-1},\nabla_X\cF(X_{t-1}+W_{t-1})}\right)\\&~~~~-\eta^2\left(\EE[\fnorm{\nabla_X\cF(X_{t-1}+W_{t-1})}^2|\cF_{t-1}]-\fnorm{\nabla_X\cF(X_{t-1}+W_{t-1})}^2\right)\big|\\
 	&\leq (1-\beta_1)^{-t}\Big[2\eta\big( (2d+1)\gamma^2 \fnorm{X_{t-1}}^2+3\fnorm{X_{t-1}}^3\fnorm{W_{t-1}}+3\fnorm{X_{t-1}}^2\fnorm{W_{t-1}}^2+\fnorm{X_{t-1}}\fnorm{W_{t-1}}^3\\&~~~~~~~~~~~+\fnorm{X_{t-1}}\fnorm{X_{t-1}}\fnorm{W_{t-1}}\big)+2\eta^2 C_1\Big]\\
 	&\leq (1-\beta_1)^{-t} \eta \phi_1,
 	\end{align*}
 	where $\phi_1= \cO( d^{1.5}).$  Then \eqref{ineq_diff} is satisfied. Directly applying Part II of Theorem \ref{thm_general}, we can get the result. Specifically, we 
 	choose
 	\begin{align*} \eta&=\min\left\{\cO\left(\frac{d^2}{d^3}\left(\log \frac{1}{\delta}\right)^{-1}\right),\cO\left(\frac{d}{d^3}\right)\right\}\\
 	&=\cO\left(\frac{1}{d^2}\left(\log \frac{1}{\delta}\right)^{-1}\right).
 	\end{align*}
 	%
 	%
 	With at least probability $1-\delta$, we have $\fnorm{X_t}^2\leq 4 d$ for all $t\leq \cO\left(\frac{1}{\eta^2}\right)$.\\
 \end{proof}

\subsection{Proof of Lemma \ref{lem_E_converge}}
\begin{proof}
	
	Let $\cF_t=\sigma\{X_\tau,\tau\leq t\}$ be the $\sigma-$field generated by past $t$ iterations. 	We first calculate the conditional expectation of $\fnorm{X_{t+1}}^2$ given $\cF_t.$ 
	Note that the update of $E_t$ can be written as follows:
	\begin{align}
	E_{t+1}=(\id-\id_{S}) X_{t+1}&=	(\id-\id_{S})\left(X_t-\eta \nabla_X\cF(X_t+W_t)\right)\nonumber\\
	&=E_t-\eta (\id-\id_{S}) \nabla_X\cF(X_t+W_t)
	\end{align}
	Then we calculate the conditional expectation of $\fnorm{E_{t+1}}^2.$
	\begin{align*}
	\EE[\fnorm{E_{t+1}}^2|\cF_t]&=\fnorm{E_t}^2-2\eta \EE[\lbr{E_t, (\id-\idu) \nabla_X\cF(X_t+W_t)}|\cF_t]\\&~~~~~+\eta^2\EE[\fnorm{ (\id-\idu) \nabla_X\cF(X_t+W_t)}^2|\cF_t].
	\end{align*}
Applying the subspace dissipative condition \eqref{pd_E}, we get the following inequality. 	
	\begin{align*}
	\EE[\fnorm{E_{t+1}}^2|\cF_t]\leq& (1-2\eta((2d+1)\gamma^2-\tnorm{\sGamma}))\fnorm{E_t}^2+\eta\frac{\tnorm{\sGamma x^*}^2}{2}\\&+\eta^2 \EE[\fnorm{ (\id-\idu) \nabla_X\cF(X_t+W_t)}^2|\cF_t]. 
	\end{align*}
	Since both $X_t$ and $W_t$ are bounded, we can  verify $\EE[\fnorm{ (\id-\idu) \nabla_X\cF(X_t+W_t)}^2|\cF_t]\leq C d^{3}.$
	Let $\beta_2=2((2d+1)\gamma^2-\tnorm{\sGamma}),$ $\alpha_2=\frac{\tnorm{\sGamma x^*}^2}{4((2d+1)\gamma^2-\tnorm{\sGamma})}$ and $\lambda_2=C \frac{d^{3}}{2((2d+1)\gamma^2-\tnorm{\sGamma})}, $  then we have 
		\begin{align*}
\EE[(\fnorm{E_{t+1}}^2-\alpha_2-\eta\lambda )|\cF_t]\leq (1-\eta\beta_2)\left(\fnorm{E_t}^2-\alpha_2-\eta\lambda\right).
\end{align*}	
Then \eqref{eq_dyn} holds for $\fnorm{E_t}^2.$ Moreover, based on the boundedness proved in Lemma \ref{lem_bound}, one can easily verify.	
\begin{align*}
\big|\fnorm{E_{t+1}}^2-\EE[\fnorm{E_{t+1}}^2|\cF_t]\big|\cI_{\{\fnorm{E_t}^2\leq 4 \alpha_2\}}\leq \eta C_2d^{2.25}\sigma^{1.5}. 
	\end{align*}
	Thus, if we take $\phi_2=C_2d^{2.25}\sigma^{1.5},$ \eqref{eq_diff} holds. By Theorem \ref{thm_general}, if we take $$\eta=\min\left\{\cO\left(\frac{\sigma}{d^3}\left(\log \frac{1}{\delta}\right)^{-1}\right),\cO\left(\frac{\sigma^2}{d^2}\right)\right\},$$ we then have with probability at least $1-\delta,$
	\begin{itemize}
	\item 	 $\fnorm{E_t}^2\leq \fnorm{E_0}^2+c_1\sqrt{d\sigma^2}\leq 1,$ for all $t's$ such that $ t\leq T=\cO(1/\eta^2);$
		\item $\fnorm{E_t}^2\leq c_1\sqrt{d\sigma^2},$ for all $t's$ such that $\tau_1\leq t\leq T=\cO(1/\eta^2),$  where
		$c_{1}$ is a constant and $$\tau_{1}={\cO}\Big(\frac{1}{\eta \sqrt{d\sigma^2}}\log\frac{\fnorm{E_0}^2}{\sqrt{d\sigma^2}}\log
		\frac{1}{\delta}\Big).$$
		\end{itemize}

\end{proof}

\subsection{Proof of Lemma \ref{lem_away}}
\begin{proof}
With our initialization, we have	$ \tnorm{r_0}^2\leq a.$ Then we  prove 	$ \tnorm{r_0}^2\leq a+\cO(\sqrt{d\sigma^2})$ for long enough time.

Recall that  $a=1-(2d+1)\gamma^2+x^{*\top}\sGamma x^*.$ By \eqref{pd_r}, the subspace dissipative condition of $r_t,$ we can upper bound the conditional expectation of $\tnorm{r_{t+1}}^2-a$ given the trajectory history.
	\begin{align*}
	\EE[\tnorm{r_{t+1}}^2-a\big|\cF_t]&=(\tnorm{r_{t}}^2-a)-2\eta \EE\left[\lbr{r_t,\nabla_X\cF(X_t+W_t)^\top x^*}\big|\cF_t\right]	\\&~~~~~~~~~~~~~~~~~~~~~+\eta^2\EE\left[\tnorm{\nabla_X\cF(X_t+W_t)^\top x^*}^2\big|\cF_t\right]\\
	&=(1-2\eta \tnorm{r_t}^2)(\tnorm{r_{t}}^2-a)-2\eta(\tnorm{E_tr_t}^2-r_t^{\top} E_t^\top\sGamma x^*)\\&~~~~~~~~~~~~~~~~~~~+\eta^2\EE\left[\tnorm{\nabla_X\cF(X_t+W_t)^\top x^*}^2\big|\cF_t\right]\\
	&\leq (1-2\eta a)(\tnorm{r_{t}}^2-a)+\eta\frac{\tnorm{\sGamma x^*}^2}{2}+\eta^2\EE\left[\tnorm{\nabla_X\cF(X_t+W_t)^\top x^*}^2\big|\cF_t\right]\\
	&\leq (1-2\eta a)(\tnorm{r_{t}}^2-a)+\eta\frac{\tnorm{\sGamma x^*}^2}{2}+\eta^2 C d^{3},
	\end{align*}
	where $C$ is a constant.
	Let $\beta_3=2 a,$ $\alpha_3=\frac{\tnorm{\sGamma x^*}^2}{4a}$ and $\lambda_3= \frac{C d^{3}}{2a}.$
	This is equivalent to 
	\begin{align*}
	\EE[\tnorm{r_{t+1}}^2-a-\alpha_3\big|\cF_t]\leq (1-\beta_3)\left(\tnorm{r_{t}}^2-a-\alpha_3-\eta \lambda_3\right).
	\end{align*}
	Then \eqref{eq_dyn} holds for $\tnorm{r_t}^2-a.$
	Denote $\cH_t=\{\forall \tau\leq t, \tnorm{r_t}^2-a\leq4\alpha_3\},$
	Then we have 
	\begin{align*}
	\EE[G_{t+1}\cI_{\cH_t}|\cF_t]\leq G_{t}\cI_{\cH_t} \leq G_t\cI_{\cH_{t-1}}.
	\end{align*}
We then  bound the difference between $\tnorm{r_{t+1}}^2\cI_{\cH_t}$ and $\EE[\tnorm{r_{t+1}}^2\cI_{\cH_t}|\cF_t].$
	\begin{align*}
\big|\tnorm{r_{t+1}}^2\cI_{\cH_t}-\EE[\tnorm{r_{t+1}}^2\cI_{\cH_t}|\cF_t]\big|\leq \eta C_3 d= \eta\phi_3.
	\end{align*}
	where $\phi_3=\cO( d).$ Thus, \eqref{eq_diff} holds for $\tnorm{r_{t}}^2-a.$ We can then apply Theorem \ref{thm_general}. Choose 
	$$\eta=\min\left\{\cO\left(\sigma^4\left(\log \frac{1}{\delta}\right)^{-1}\right),\cO\left(\frac{\sigma^2}{d^2}\right)\right\},$$ 
%
%
%
then  with probability $1-\delta$, we have $\norm{r_t}_2^2\leq a+4\alpha_3$ for all t's such that $ t\leq\cO(\eta^{-2}).$

	We next prove \eqref{eq_r1} .	
Suppose there exists some time $t$ such that $\norm{r_t}^2\geq a-2/3,$ our following analysis will show that the algorithm will stay in  the region such that $\norm{r_t}^2\geq a-{2}/{3},$ for long enough time. Then we can move to Lemma \ref{lem_r_converge}.  Suppose such $t$ does not exists, i.e., $\norm{r_t}^2\leq a-2/3,$ for all $t's.$ Then we have the following inequality.
	\begin{align*}
\EE[\tnorm{r_{t+1}}^2\big|\cF_t]&=\tnorm{r_{t}}^2-2\eta \EE\left[\lbr{r_t,\nabla_X\cF(X_t+W_t)^\top x^*}\big|\cF_t\right]	+\eta^2\EE\left[\tnorm{\nabla_X\cF(X_t+W_t)^\top x^*}^2\big|\cF_t\right]\\
&=\left(1-2\eta (\tnorm{r_t}^2-a)\right)\tnorm{r_{t}}^2-2\eta(\tnorm{E_tr_t}^2-r_t^{\top} E_t^\top\sGamma x^*)\\
&\hspace{+2in}+\eta^2\EE\left[\tnorm{\nabla_X\cF(X_t+W_t)^\top x^*}^2\big|\cF_t\right]\\
&\geq \left(1-\eta (2\tnorm{E_t}^2)\right)\tnorm{r_{t}}^2+\eta(2(a-\tnorm{r_t}^2)\tnorm{r_t}^2-2\tnorm{r_t}\tnorm{E^\top_t\sGamma x^*})\\
&\geq \left(1-\eta \left(2+{2.5-2}\right)\right)\tnorm{r_{t}}^2+\eta\left(2(a-\tnorm{r_t}^2)\tnorm{r_t}^2-2\tnorm{E^\top_t\sGamma x^*}^2\right)\\
&\geq  \left(1-2.5\eta \right)\tnorm{r_{t}}^2+\eta\left(3\tnorm{r_t}^2-4r^2\tnorm{\sGamma x^*}^2\right), 
\end{align*}
	where $r^2 = \frac{c_1}{2}\sqrt{d\sigma^2}$. Let $\cE_t=\left\{\tnorm{r_\tau}^2 \geq r^2\tnorm{\sGamma x^*}, \forall \tau\leq t\right\}.$  Then we have 
	\begin{align*}
	&\EE\left[(1-2.5\eta )^{-t-1}\left(\tnorm{r_{t+1}}^2-\frac{3-4\tnorm{\sGamma x^*}}{2.5}r^2\tnorm{\sGamma x^*}\right)\cI_{\cE_t}]\big|\cF_t\right]\\&~~~~~~~~~~~~~~~~~~~~~~~~~\geq (1-2.5\eta )^{-t}\left(\tnorm{r_{t}}^2-\frac{3-4\tnorm{\sGamma x^*}}{2.5}r^2\tnorm{\sGamma x^*}\right)\cI_{\cE_t}
	\\&~~~~~~~~~~~~~~~~~~~~~~~~~\geq (1-2.5\eta )^{-t}\left(\tnorm{r_{t}}^2-\frac{3-4\tnorm{\sGamma x^*}}{2.5}r^2\tnorm{\sGamma x^*}\right)\cI_{\cE_{t-1}}.
	\end{align*}
The last inequality comes from the fact $\frac{3-4\tnorm{\sGamma x^*}}{2.5}\geq 1.$ The above inequality actually shows that 
$$G_t=(1-2.5\eta )^{-t}\left(\tnorm{r_{t}}^2-\frac{3-4\tnorm{\sGamma x^*}}{2.5}r^2\tnorm{\sGamma x^*}\right)\cI_{\cE_{t-1}}.$$ is a submartingale. Following the same proof of Part II of Theorem \ref{thm_general}, we can show that with our choice of small $\eta,$ with high probability, $ \tnorm{r_{t}}^2\geq r^2 \tnorm{\sGamma x^*}\geq \tnorm{\sGamma x^*}^2.$
\end{proof}

\subsection{Proof of Lemma \ref{lem_r_converge}}
\begin{proof}
We first show that there must exist some $\tau_{21}$ such that $\tnorm{r_{t}}^2>\frac{a}{3}.$ We first have the following inequality:
	\begin{align*}
\EE[\tnorm{r_{t+1}}^2\big|\cF_t]&=\tnorm{r_{t}}^2-2\eta \EE\left[\lbr{r_t,\nabla_X\cF(X_t+W_t)^\top x^*}\big|\cF_t\right]	+\eta^2\EE\left[\tnorm{\nabla_X\cF(X_t+W_t)^\top x^*}^2\big|\cF_t\right]\\
&=\left(1-2\eta (\tnorm{r_t}^2-a)\right)\tnorm{r_{t}}^2-2\eta(\tnorm{E_tr_t}^2-r_t^{\top} E_t^\top\sGamma x^*)\\&\hspace{+2.5in}+\eta^2\EE\left[\tnorm{\nabla_X\cF(X_t+W_t)^\top x^*}^2\big|\cF_t\right]\\
&\geq \left(1-2\eta (\tnorm{r_t}^2-a)\right)\tnorm{r_{t}}^2-2\eta(\frac{3}{2}\fnorm{E_t}^2\tnorm{r_t}^2+\frac{\tnorm{\sGamma x^{*}}^2}{2})\\&\hspace{+2.5in}+\eta^2\EE\left[\tnorm{\nabla_X\cF(X_t+W_t)^\top x^*}^2\big|\cF_t\right]\\
&\geq \left(1-2\eta (\tnorm{r_t}^2-a+\frac{3}{2}c_1\epsilon )\right)\tnorm{r_{t}}^2-\eta{\tnorm{\sGamma x^{*}}^2}.
\end{align*}
Denote $\cE_t=\{\forall \tau\leq t, \tnorm{r_{\tau}}^2\leq \frac{a}{3}\}.$ Then we have 
	\begin{align*}
\EE[\tnorm{r_{t+1}}^2\cI_{\cE_t}\big|\cF_t]
&\geq \left(1-2\eta (-\frac{2}{3}a+\frac{3}{2}c_1\epsilon )\right)\tnorm{r_{t}}^2\cI_{\cE_t}-\eta{\tnorm{\sGamma x^{*}}^2}\cI_{\cE_t}.
\end{align*}
Let $G_t= \left(1+\eta (\frac{4}{3}a-3c_1\epsilon )\right)^{-t}\left(\tnorm{r_{t}}^2-\frac{\tnorm{\sGamma x^{*}}^2}{\frac{4}{3}a-3c_1\epsilon}\right).$
Thus, we have $$\EE\left[G_{t+1}\cI_{\cE_t}\big|\cF_t\right]\geq G_{t}\cI_{\cE_t}\geq G_{t}\cI_{\cE_{t-1}}. $$
The last inequality must hold, otherwise we have found a $t$ such that $\tnorm{r_{t}}^2>\frac{a}{3}.$ We have constructed a sub-martingale sequence.
Following similar lines to our previous proof, with probability at least $1-\delta,$ there exists $t\leq \tau_{21}= \frac{1}{a\eta}\log \frac{4a}{d\sigma^2}\log\frac{1}{\delta},$ such that  $\tnorm{r_{t}}^2>\frac{a}{3}.$

Next, we show that the solution trajectory will stay in this region $\{\tnorm{r_t}^2>\frac{a}{3}\}$. Let $\cE_t=\{\forall \tau\leq t, \tnorm{r_\tau}^2>\frac{a}{3}\}.$
\begin{align*}
\EE[a-\tnorm{r_{t+1}}^2\big|\cF_t]&=(a-\tnorm{r_{t}}^2)+2\eta \EE\left[\lbr{r_t,\nabla_X\cF(X_t+W_t)^\top x^*}\big|\cF_t\right]	\\&\hspace{+2.5in}-\eta^2\EE\left[\tnorm{\nabla_X\cF(X_t+W_t)^\top x^*}^2\big|\cF_t\right]\\
&=(1-2\eta \tnorm{r_t}^2)(a-\tnorm{r_{t}}^2)+2\eta(\tnorm{E_tr_t}^2-r_t^{\top} E_t^\top\sGamma x^*)\\&\hspace{+2.5in}-\eta^2\EE\left[\tnorm{\nabla_X\cF(X_t+W_t)^\top x^*}^2\big|\cF_t\right].
\end{align*}
Let $a'=a+4\alpha_3,$ where $\alpha_3= \frac{\tnorm{\sGamma x^*}^2}{4a}$ comes from the proof of Lemma \ref{lem_away}, and thus $\tnorm{r_t}^2\leq a'.$ Then the above equality is equivalent to the following:
\begin{align*}
\EE[a'-\tnorm{r_{t+1}}^2\big|\cF_t]&=(1-2\eta \tnorm{r_t}^2)(a'-\tnorm{r_{t}}^2)+2\eta (a'-a) \tnorm{r_t}^2\\&~~~~~~~~~~+2\eta(\tnorm{E_tr_t}^2-r_t^{\top} E_t^\top\sGamma x^*)-\eta^2\EE\left[\tnorm{\nabla_X\cF(X_t+W_t)^\top x^*}^2\big|\cF_t\right]\\
&\leq (1-2\eta \tnorm{r_t}^2)(a'-\tnorm{r_{t}}^2)+2\eta (a'-a) a'+2\eta(\tnorm{E_tr_t}^2-r_t^{\top} E_t^\top\sGamma x^*).
\end{align*}
We further have
\begin{align*}
\EE[a'-\tnorm{r_{t+1}}^2\cI_{\cE_t}\big|\cF_t]
&\leq \left(1-\eta\frac{2}{3} a\right)(a'-\tnorm{r_{t}}^2)\cI_{\cE_t}+C_4\sqrt{d\sigma^2}\cI_{\cE_t},
\end{align*}
which is equivalent to the following equations.
\begin{align*}
\EE[(a'-\tnorm{r_{t+1}}^2-C_4\sqrt{d\sigma^2})\cI_{\cE_t}\big|\cF_t]
\leq  \left(1-\eta\frac{2}{3} a\right)\left(a'-\tnorm{r_{t}}^2-C_4a\sqrt{d\sigma^2}\right)\cI_{\cE_{t-1}}.
\end{align*}
Then we can construct a supermartingale $G_t\cI_{\cE_{t-1}}=\frac{1}{ (1-\eta\frac{2}{3} a)^t}(a'-\tnorm{r_{t}}^2-C_4\sqrt{d\sigma^2})\cI_{\cE_{t-1}}.$ Applying Theorem \ref{thm_general}, one can show with probability $1-\delta$, we have $a'-\norm{r_t}_2^2\leq 4\frac{a'-\frac{a}{3}}{4}$ or equivalently $\norm{r_t}_2^2\geq \frac{a}{3}$ for all t's such that $\tau_{21}\leq t\leq{\cO}(\eta^{-2}).$ Then the following inequality always holds.
\begin{align*}
\EE[(a'-\tnorm{r_{t+1}}^2-C_4\sqrt{d\sigma^2})\big|\cF_t]
\leq  \left(1-\eta\frac{2}{3} a\right)\left(a'-\tnorm{r_{t}}^2-C_4\sqrt{d\sigma^2}\right).
\end{align*}
	Following similar lines to the proof of $\norm{r_t}_2^2\leq a+4\alpha_3$, one can show with probability $1-\delta$, we have $\norm{r_t}_2^2\geq a'-C_5 \sqrt{d\sigma^2}$ for all t's such that $\tau_{22}\leq t\leq{\cO}(\eta^{-2})$, where
 $\tau_{22}={\cO}\Big(\frac{1}{\eta}\log\frac{1}{d\sigma^2}\log
	\frac{1}{\delta}\Big).$ Take $\tau_2=\tau_{21}+\tau_{22},$ we have when $t\geq \tau_2,$
	$$ a'-C_5 \sqrt{d\sigma^2}\leq \norm{r_t}_2^2\leq a+4\alpha_3.$$
	Therefore there exists some constant $c_2>0$ such that when $t\geq \tau_2,$
		$$  |\norm{r_t}_2^2-1|\leq c_2\sqrt{d\sigma^2}.$$
\end{proof}

\section{Proof of Lemma \ref{lem_er_converge}}
\begin{proof}
	Note that we can refine the upper bound of the norm of $X_t$ as follows: $\fnorm{X_t}^2=\fnorm{E_t}^2+\tnorm{r_t}^2\leq (c1+c2)\sqrt {d\sigma^2}.$ 
			We first write down the update of $E_tr_t:$
	\begin{align*}
	E_{t+1}r_{t+1}=E_tr_t-\eta E_t \nabla_{r}\cF(X_t+W_t)-\eta \nabla_{E}\cF(X_t+W_t)r_t+\eta^2 \nabla_{E}\cF(X_t+W_t)\nabla_{r}\cF(X_t+W_t).
	\end{align*}
 For notational simplicity, denote $D_{1,t}= \nabla_{E}\cF(X_t+W_t)\nabla_{r}\cF(X_t+W_t).$ By simple calculation, we know that $\tnorm{D_{1,t}}$ is at most $\cO(d\sigma^2).$	Then the update of the squared norm of  $E_tr_t$ is as follows:
	\begin{align*}
\tnorm{E_{t+1}r_{t+1}}^2&=\tnorm{E_tr_t}^2-2\eta (E_tr_t)^\top  E_t \nabla_{r}\cF(X_t+W_t)-2\eta (E_tr_t)^\top\nabla_{E}\cF(X_t+W_t)r_t\\
&~~~~+\eta^2 \left (\tnorm{E_t \nabla_{r}\cF(X_t+W_t)}^2+\tnorm{\nabla_{E}\cF(X_t+W_t)r_t}^2+2(E_tr_t)^\top D_{1,t}\right)\\
&~~~~-2\eta^3 D_{1,t}^{\top}\left( E_t \nabla_{r}\cF(X_t+W_t)+ \nabla_{E}\cF(X_t+W_t)r_t\right)+\eta^4 \tnorm{D_{1,t}}^2\\
&=\tnorm{E_tr_t}^2-2\eta (E_tr_t)^\top  E_t \nabla_{r}\cF(X_t+W_t)-2\eta (E_tr_t)^\top\nabla_{E}\cF(X_t+W_t)r_t+\eta^2 D_{2,t},
	\end{align*}
	where 
	\begin{align*}
D_{2,t}=& \left (\tnorm{E_t \nabla_{r}\cF(X_t+W_t)}^2+\tnorm{\nabla_{E}\cF(X_t+W_t)r_t}^2+2(E_tr_t)^\top D_{1,t}\right)\\
&~~-2\eta D_{1,t}^{\top}\left( E_t \nabla_{r}\cF(X_t+W_t)+ \nabla_{E}\cF(X_t+W_t)r_t\right)+\eta^2 \tnorm{D_{1,t}}^2.
	\end{align*}
By simple calculation, we know $D_{2,t}$ is at most $\cO(1).$ Thus,  the last three terms is $\eta^2 {D_{2,t}}\leq C_6\eta^2,$ and the update is dominated by the $\cO(\eta)$ terms. We next calculate the $\cO(\eta)$ terms as follows
\begin{align*}
&\EE[(E_tr_t)^\top\nabla_{E}\cF(X_t+W_t)r_t|\cF_t]\\
=&(E_tr_t)^\top(\id-\idu)\left((X_tX_t^\top-Y^*)X_t-\sGamma X_t+(2d+1)\gamma^2 X_t \right)r_t\\
=&r_t^\top E_t^\top\Big((x^*r_t^\top+E_t)(E_t^\top E_tr_t+r_tr_t^\top r_t)-x^*r_t^\top r_t-\sGamma E_tr_t\\&\hspace{+2in}-\sGamma x^*r_t^\top r_t+(2d+1)\gamma^2E_tr_t+(2d+1)\gamma^2 x^*r_t^\top r_t \Big)\\
=&\left(\tnorm{r_t}^2+r_t^\top E_t^\top x^*+(2d+1)\gamma^2\right)\tnorm{E_tr_t}^2-r_t^\top E_t^\top\sGamma E_tr_t\\
&~~~~~+\left(\tnorm{r_t}^4-\tnorm{r_t}^2+(2d+1)\gamma^2\tnorm{r_t}^2\right)r_t^\top E_t^\top x^*+\tnorm{E_t^\top E_tr_t}^2+\tnorm{r_t}^2r_t^\top E_t^\top\sGamma x^*\\
\geq& \left(\frac{3}{4}\tnorm{r_t}^2-\tnorm{\sGamma}+r_t^\top E_t^\top x^*+(2d+1)\gamma^2\right)\tnorm{E_tr_t}^2+\tnorm{r_t}^2\left(\tnorm{r_t}^2-1+(2d+1)\gamma^2\right)r_t^\top E_t^\top x^*\\&\hspace{+2.5in}+\tnorm{r_t}^2\left(\frac{1}{4}\tnorm{E_tr_t}^2-r_t^\top E_t^\top\sGamma x^*\right)\\
\geq& \left(\frac{3}{4}\tnorm{r_t}^2+\tnorm{\sGamma}+r_t^\top E_t^\top x^*+(2d+1)\gamma^2\right)\tnorm{E_tr_t}^2\\&\hspace{+2in}+\tnorm{r_t}^2\left(\tnorm{r_t}^2-1+(2d+1)\gamma^2\right)r_t^\top E_t^\top x^*-\frac{1}{2}\tnorm{\sGamma x^*}^2,
\end{align*}
and 
\begin{align*}
&\EE[(E_tr_t)^\top  E_t \nabla_{r}\cF(X_t+W_t)|\cF_t]\\=&r_t^\top E_t^\top E_t \left((X_t^\top X_tX_t^\top-Y^*)x^*-X_t^\top \sGamma x^* +(2d+1)\gamma^2 X_t^\top x^* \right)\\
=&\tnorm{E_t^\top E_tr_t}^2+\left(\tnorm{r_t}^2-1-x^{*\top} \sGamma x^*+(2d+1)\gamma^2\right)\tnorm{E_tr_t}^2-r_t^\top E_t^\top E_tE_t^\top \sGamma x^*\\
\geq &\left(\tnorm{r_t}^2-a\right)\tnorm{E_tr_t}^2-r_t^\top E_t^\top E_tE_t^\top \sGamma x^*.
\end{align*}
Combine the above two inequalities together and we have:
\begin{align*}
\EE[\tnorm{E_{t+1}r_{t+1}}^2|\cF_t]&\leq \left(1-2\eta\left(\frac{7}{4}\tnorm{r_t}^2-a-\tnorm{\sGamma}+r_t^\top E_t^\top x^*+(2d+1)\gamma^2\right)\right)\tnorm{E_tr_t}^2\\
   &~~~~~-2\eta\Big(\tnorm{r_t}^2\left(\tnorm{r_t}^2-1+(2d+1)\gamma^2\right)r_t^\top E_t^\top x^*\\&\hspace{+1in}-\frac{1}{2}\tnorm{\sGamma x^*}^2-r_t^\top E_t^\top E_tE_t^\top \sGamma x^*\Big)+\eta^2 D_{2,t} \\
   &\leq (1-\eta)\tnorm{E_tr_t}^2+\eta C_7 d\sigma^2+C_6\eta^2.
\end{align*}
Let $\alpha_4=C_7d\sigma^2,$ $\lambda_4=C_6$ and $\beta=1$, then $\eqref{eq_dyn}$ holds.  Moreover, one can also check $|\tnorm{E_tr_t}^2-\EE[\tnorm{E_tr_t}^2|\cF_{t-1}]\leq \eta C_8,$ where $C_8$ is some constant.   Then we can apply Theorem \ref{thm_general}. Choose 
$$\eta=\cO\left(d\sigma^2\left(\log \frac{1}{\delta}\right)^{-1}\right),$$  then with probability at least $1-\delta,$ there exists some constant $c_3>0$ such that
$$\tnorm{E_{t+1}r_{t+1}}^2\leq c_3d\sigma^2,$$
for all $t's$ such that $\tau_3\leq t\leq \cO(\frac{1}{\eta^2}),$ where $\tau_3=\cO(\frac{1}{\eta}\log\frac{1}{d\sigma^2}\log\frac{1}{\delta}).$ 
\end{proof}

\subsection{Proof of Lemma \ref{lem_noise}}
\begin{proof}
	Note that the Frobenius norm of $\Gamma$ can be written as a sum of $d^2$ squared subGaussian random variable: $\fnorm{\Gamma}^2=\sum_{i,j} \Gamma_{ij}^2.$ Since $\Gamma_{i,j}$ is subGaussian, $\Gamma_{i,j}^2$ is sub-exponential. Then we have the following concentration inequality.
	\begin{align*}
	\PP\left(\big|\frac{\fnorm{\Gamma}}{d}-\sigma\big|\geq t\right)\leq 2\exp\left(-\frac{d^2t^2}{2C\sigma^2}\right),
	\end{align*}
	for any $t>0.$	Take $t=\frac{\sigma}{d}\sqrt{2C\log\frac{8}{\delta}},$ we have with probability at least $1-\frac{\delta}{4},$ we have
	$$\fnorm{\Gamma}\leq d\sigma+\sigma\sqrt{2C\log\frac{8}{\delta}}.  $$
	Then we have $$\fnorm{\sGamma}\leq \frac{1}{2}(\fnorm{\Gamma}+\fnorm{\Gamma^\top})=\fnorm{\Gamma}\leq d\sigma+\sigma\sqrt{2C\log\frac{8}{\delta}}.$$
	Moreover, since $\tnorm{x^*}=1,$ we have 
	\begin{align*}
	\PP\left(\big|\frac{\tnorm{\Gamma x^*}}{\sqrt{d}}-\sigma\big|\geq t\right)\leq 2\exp\left(-\frac{dt^2}{2C\sigma^2}\right),
	\end{align*}
	\begin{align*}
	\PP\left(\big|\frac{\tnorm{\Gamma^\top x^*}^2}{\sqrt{d}}-\sigma\big|\geq t\right)\leq 2\exp\left(-\frac{dt^2}{2C\sigma^2}\right).
	\end{align*}
	Take $t=\frac{\sigma}{\sqrt{d}}\sqrt{2c\log\frac{8}{\delta}},$ we have with probability at least $1-\frac{\delta}{4},$ we have
	$$\tnorm{\Gamma x^*}\leq \sqrt{d}\sigma+\sigma\sqrt{2C\log\frac{8}{\delta}},  $$
	$$\tnorm{\Gamma^\top x^*}\leq \sqrt{d}\sigma+\sigma\sqrt{2C\log\frac{8}{\delta}}. $$
	Then we have $$\tnorm{\sGamma x^*}\leq \frac{1}{2}(\fnorm{\Gamma x^*}+\fnorm{\Gamma^\top x^*})\leq \sqrt{d}\sigma+\sigma\sqrt{2C\log\frac{8}{\delta}}.$$
	
	By Theorem 4.4.5 in \cite{vershynin2019high}, we have for any $t>0,$
	$$\tnorm{\Gamma}\leq C\sigma(2\sqrt{d}+t),$$
	with probability at least $1-2\exp(-t^2),$ where $C$ is some absolute constant.  Take $t=\sqrt{\log \frac{2}{\delta}},$ we have with probability at least $1-\delta,$
	$$\tnorm{\Gamma}\leq C\sigma\left(2\sqrt{d}+\sqrt{\log \frac{2}{\delta}}\right).$$
	Then we have $$\tnorm{\sGamma}\leq \frac{1}{2}(\tnorm{\Gamma}+\tnorm{\Gamma^\top})=\tnorm{\Gamma}\leq C\sigma\left(2\sqrt{d}+\sqrt{\log \frac{2}{\delta}}\right).$$
	Take $\delta=\cO(\exp(-d))$ and we prove the result.
\end{proof}
\subsection{Proof of Lemma \ref{lem_initial}}
\begin{proof}
	Note that our initialization can be rewritten as $X_0=r x'_0{x'}_0^{\top},$ where $r^2\sim\mathrm{UNIF}[0,1]$ and $x'_0\sim \mathrm{UNIF}(\SSS(1)).$ Then 
	\begin{align*}
	\tnorm{r_0}^2=\tnorm{X_0^\top x^*}^2=r^2 ({x'}_0^{\top} x^*)^2=r^2\cos(\angle(x'_0,x^*))^2.
	\end{align*}
	Note that the probability $ \tnorm{r_0}^2\geq C^2 d\sigma^2$ can then be bounded as follows.
	\begin{align*}
	\PP\left(r^2\cos(\angle(x'_0,x^*))^2\geq  C^2 d\sigma^2\right)&\geq \PP\left(r^2\geq C\sqrt{d\sigma^2}, \cos(\angle(x'_0,x^*))^2\geq  C\sqrt{d\sigma^2} \right)\\
	&\geq \PP\left(r^2\geq C\sqrt{d\sigma^2}\right)+\PP\left( \cos(\angle(x'_0,x^*))^2\geq  C\sqrt{d\sigma^2} \right)-1\\
	&=1-  \PP\left(r^2\leq C\sqrt{d\sigma^2}\right)-\PP\left( \cos(\angle(x'_0,x^*))^2\leq { C\sqrt{d\sigma^2}} \right)\\
	&= 1- \PP\left(r^2\leq  C\sqrt{d\sigma^2}\right)-4\PP\left (\arccos(\sqrt{  C\sqrt{d\sigma^2}})\leq\theta \leq \frac{\pi}{2}\right)\\
	&=1-\cO\left(\frac{1}{d^{0.25}}\right),
	\end{align*}
	where $\theta\sim\mathrm{UNIF}[0,\pi/2].$ That is with high probability, we have	$ \tnorm{r_0}^2\geq \tnorm{\sGamma x^*}^2.$
	
	Moreover, $\PP(\fnorm{X_0}^2\leq 1- C_1\sqrt{d\sigma^2})=\PP(r^2\leq 1- C_1\sqrt{d\sigma^2})=1- C_1\sqrt{d\sigma^2}=1-\cO(\frac{1}{d^{0.25}}).$ We finish the proof.
\end{proof}

\end{document}